\documentclass[conference]{IEEEtran}

\IEEEoverridecommandlockouts
\usepackage{cite}
\usepackage{amsmath,amssymb,amsfonts}
\usepackage{algorithmic}
\usepackage{graphicx}
\usepackage{textcomp}
\usepackage{xcolor}
\usepackage{adjustbox}
\usepackage{booktabs}
\usepackage{threeparttable}
 \usepackage{subcaption}
 \usepackage{algorithm}
 \usepackage{makecell}
\usepackage{algorithmic}
\usepackage{url}
\newcolumntype{C}[1]{>{\centering\arraybackslash}m{#1}} 
\newcolumntype{L}[1]{>{\raggedright\arraybackslash}m{#1}} 
\def\BibTeX{{\rm B\kern-.05em{\sc i\kern-.025em b}\kern-.08em
    T\kern-.1667em\lower.7ex\hbox{E}\kern-.125emX}}
\begin{document}

\title{\title{ALPINE: Closed-Loop Adaptive Privacy Budget Allocation for Mobile Edge Crowdsensing}}

\author{
Guanjie Cheng$^{1}$, Siyang Liu$^{1}$, Xinkui Zhao$^{1}$, Yishan Chen$^{2*}$, 
Junqin Huang$^{3}$, Linghe Kong$^{3}$, Shiguang Deng$^{1}$ \\
$^{1}$Zhejiang University, Zhejiang, China \\
$^{2}$Jiangxi University of Science and Technology, Jiangxi, China \\
$^{3}$Shanghai Jiao Tong University, Shanghai, China \\
Email: \{chengguanjie, liusiyang0926, zhaoxinkui, dengsg\}@zju.edu.cn \\
chenys@just.edu.cn, \{junqin.huang, linghe.kong\}@sjtu.edu.cn
}
\maketitle

\begin{abstract}
Mobile edge crowdsensing (MECS) enables large-scale real-time sensing services, but its continuous data collection and transmission pipeline exposes terminal devices to dynamic privacy risks.
Existing privacy protection schemes in MECS typically rely on static configurations or coarse-grained adaptation, making them difficult to balance privacy, data utility, and device overhead under changing channel conditions, data sensitivity, and resource availability. To address this problem, we propose ALPINE, a lightweight closed-loop framework for adaptive privacy budget allocation in MECS. 
ALPINE performs multi-dimensional risk perception on terminal devices by jointly modeling channel, semantic, contextual, and resource risks, and maps the resulting risk state to a privacy budget through an offline-trained TD3 policy. 
The selected budget is then used to drive local differential privacy perturbation before data transmission, while edge-side privacy–utility evaluation provides feedback for policy switching and periodic refinement. 
In this way, ALPINE forms a terminal–edge collaborative control loop that enables real-time, risk-adaptive privacy protection with low online overhead. 
Extensive experiments on multiple real-world datasets show that ALPINE achieves a better privacy–utility trade-off than representative baselines, reduces the effectiveness of membership inference, property inference, and reconstruction attacks, and preserves robust downstream task performance under dynamic risk conditions. 
Prototype deployment further demonstrates that ALPINE introduces only modest runtime overhead on resource-constrained devices.
\end{abstract}


\section{Introduction}
With the rapid development of the Internet of Things (IoT), mobile edge crowdsensing (MECS) has become a promising distributed service paradigm for real-time sensing, data analytics, and context-aware applications in urban and industrial environments \cite{luan2025stability}. 
In a typical MECS workflow, large numbers of terminal devices continuously collect contextual data, edge nodes provide near-source processing and rapid response, and cloud servers support global coordination and service-level analytics. 
Such terminal–edge–cloud collaboration enables a wide range of delay-sensitive services, including smart transportation, urban monitoring, environmental surveillance, and personalized mobile applications. 
For example, sensors in public transport and traffic infrastructures can collaboratively generate real-time traffic and environmental data streams for traffic optimization and city-management services \cite{wang2025crowdradar}, while wearable and mobile devices can support environmental monitoring and pollution-control services through continuous sensing and edge-assisted analysis \cite{zhang2025fueling}. 
As an evolution of mobile crowdsensing, MECS integrates distributed sensing and edge intelligence to support low-latency, privacy-aware, and resource-constrained service provisioning \cite{10579548}.

However, this edge-assisted multi-terminal workflow also makes privacy protection significantly more challenging.
Sensitive data are generated and transmitted by heterogeneous terminal devices over open wireless links, often under strict real-time constraints. 
To reduce exposure during transmission and meet regulatory requirements, privacy-preserving operations increasingly need to be executed at or near the terminal devices \cite{jia2025transparent}. 
Yet terminal devices typically have limited computation and storage resources, and many integrated sensing and communication (ISAC) applications demand low-latency and reliable wireless service delivery \cite{zhang2026integrated}.  
The dynamic, open nature of wireless networks further exposes transmissions to eavesdropping, interference and man-in-the-middle attacks, undermining confidentiality \cite{fereidouni2025iot}.
Once data are intercepted or accessed, adversaries can further exploit techniques such as membership inference and property inference to extract sensitive personal information. 
Traditional static privacy-preserving methods, such as k-anonymity, t-closeness, rule-based generalization and suppression, and fixed-budget differential privacy (DP), are difficult to adapt to risk variations in highly dynamic edge environments. 
As a result, they often lead to either excessive perturbation and utility degradation or insufficient privacy protection \cite{zhu2025dynamic}.

\begin{figure}
  \centering
  \includegraphics[width=\linewidth]{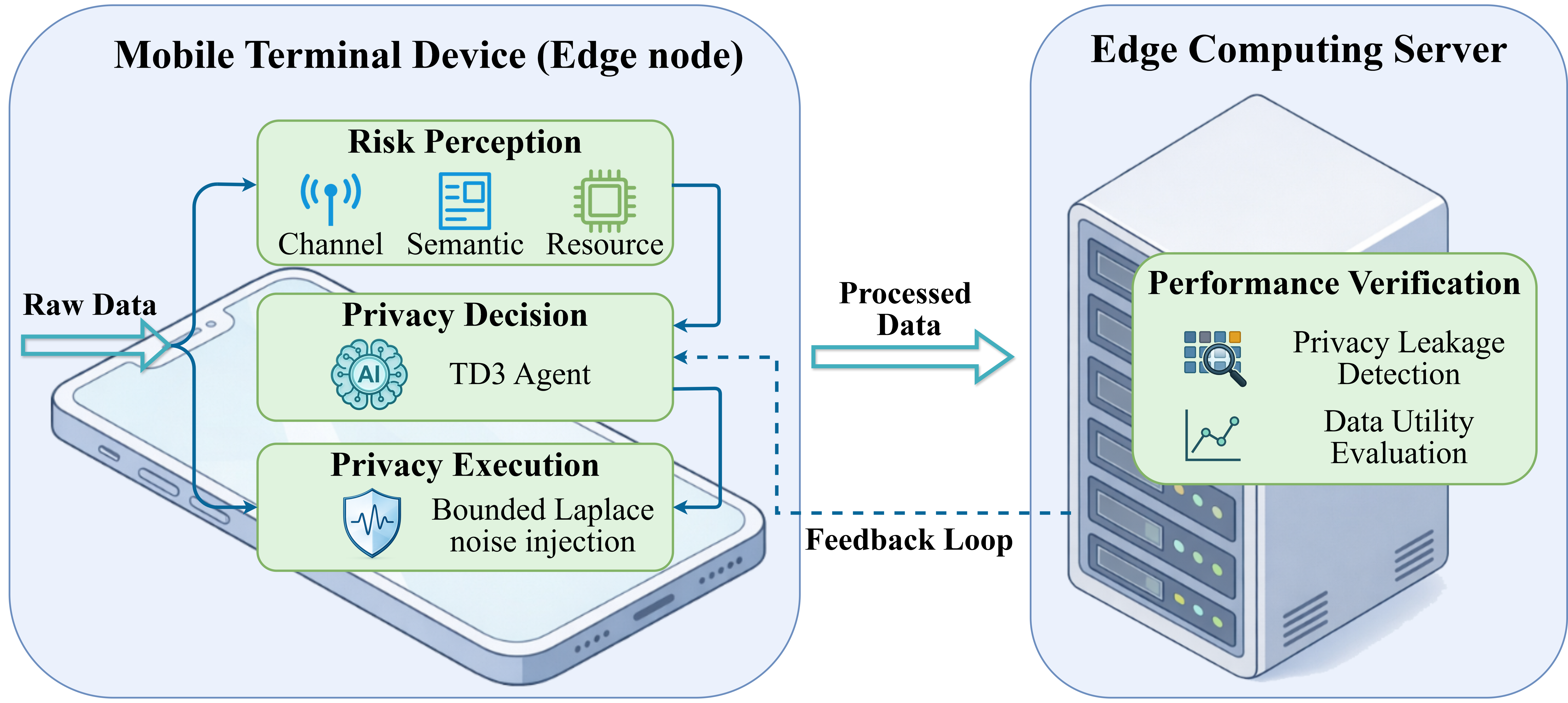}
  \caption{Overview of ALPINE. }
  \label{fig:xx}
\end{figure}

To cope with evolving threats in MECS services, recent research has begun to explore adaptive privacy protection that adjusts protection strength according to environmental risk, data sensitivity, and service context  \cite{he2025addressing}. To better accommodate instantaneous variations in network conditions and data distributions, online learning techniques have also been introduced into privacy-preserving strategies, enabling more flexible adjustment of protection intensity in real-world scenarios \cite{kiani2025differentially}. However, the inherently resource-constrained nature of terminal devices imposes stringent requirements on computational efficiency and deployment overhead \cite{zhu2025scope}. Therefore, a key challenge remains open: how to perform lightweight and real-time privacy budget allocation on terminal devices, so that protection strength can adapt quickly to environmental changes and service requirements without introducing excessive latency or energy cost.

In this study, we propose \textbf{ALPINE}, a closed-loop adaptive privacy budget allocation framework for MECS. As shown in Figure 1, ALPINE forms a terminal–edge collaborative control-loop that dynamically coordinates privacy protection, data utility, and system overhead in MECS environments. 
On the terminal side, ALPINE continuously monitors channel risk, semantic risk, and device resource status, and employs a TD3-based agent to allocate privacy budgets dynamically before differential privacy noise is applied. 
On the edge side, ALPINE leverages privacy–utility feedback from downstream evaluation to support online policy switching and periodic offline refinement. In this way, ALPINE enables practical and low-overhead adaptive privacy control for large-scale heterogeneous edge environments under stringent resource constraints.

\begin{itemize}
    \item \textbf{Closed-loop adaptive privacy budget allocation framework for MECS.} ALPINE introduces a dynamic control cycle in which a TD3 agent allocates privacy budgets in response to real-time risk, guided by a multi-objective reward function that jointly considers privacy gain, utility loss, and energy cost. The loop spans terminal-side risk perception, budget execution, and edge-side feedback, supports feedback-driven policy switching and periodic offline retraining under varying environmental conditions.
    \item \textbf{Lightweight on-device privacy adaptation mechanism.} All key models, including a block-structured lightweight model LightAE for channel-risk detection and a TD3 agent for privacy-budget allocation, are trained offline, while the online stage performs only lightweight inference. This design ensures real-time performance and low deployment overhead on resource-constrained devices.
    \item \textbf{Privacy-preserving analysis and extensive empirical evaluation.} We analyze the privacy properties of the proposed mechanism and conduct extensive experiments on real-world datasets to validate the effectiveness. Results show that ALPINE can effectively mitigate representative privacy attacks while maintaining a favorable privacy–utility trade-off in dynamic environments.
\end{itemize}

Meanwhile, we provide the full code at \url{https://anonymous.4open.science/r/ALPINE-2061/} to support reproducibility.

\section{Related Work}

\textbf{Static Privacy Protection Mechanisms.} 
Static privacy protection mechanisms typically rely on fixed perturbation strengths or predefined anonymization rules that remain unchanged across runtime conditions. In MECS scenarios, such designs are attractive because of their low implementation complexity and predictable deployment cost. Representative studies include fixed-budget Laplace perturbation for privacy-preserving task allocation \cite{ZHANG2024103464}, k-anonymity-based anonymous location matrices for location protection \cite{10945149}, and Square Wave randomization for stream-wise perturbation \cite{11017586}. 
However, fixed settings are brittle under evolving adversaries, resource budgets, and task requirements, often yielding either unnecessary utility loss or insufficient protection.
This makes them unsuitable for fine-grained privacy budget allocation in dynamic MECS environments.

\textbf{Dynamic Adaptive Privacy Protection Mechanisms.}
Dynamic privacy protection mechanisms adjust protection strength according to contextual signals such as environmental risk, data sensitivity, or resource availability. These approaches enable privacy parameters or protection mechanisms to be updated online in response to temporal dynamics and changing operating conditions. For example, Shuai et al. \cite{shuai2022r} developed a risk-adaptive differential privacy scheme for IIoT data transmission, Pan and Feng \cite{pan2023differential} studied dynamic budget allocation under zero-concentrated differential privacy, Chen et al. \cite{CHEN2024110613} proposed an online quality-aware privacy-preserving task allocation method, and Tang et al. \cite{10963891} introduced an adaptive credibility-aware privacy-preserving data collection scheme for zero-trust crowdsensing. 
Although these studies move beyond static protection, most of them rely on relatively coarse adaptation signals or optimize only a limited aspect of the privacy process. They generally do not provide a terminal–edge closed loop in dynamic MECS settings, resulting in coarse-grained budget allocation and limited robustness.

\textbf{Collaborative and Lightweight Privacy Protection Mechanisms.}
A parallel line of work studies collaborative system architectures and lightweight mechanisms for privacy-preserving edge intelligence. Cloud–edge collaboration has been widely explored to reduce end-to-end latency and keep raw data closer to devices, while sharing only intermediate updates or compressed information for privacy-aware analytics \cite{wang2020privacy}. Related studies include adaptive differential privacy for federated learning via clipping and regularization \cite{hu2024does}, as well as client–server mobile crowdsensing designs for efficient privacy-preserving analytics \cite{11045510, 11145334}. In addition, lightweight deployment strategies attempt to reduce device-side overhead through computation offloading or system-level support mechanisms, such as delegating security functions to edge servers \cite{sun2024two} or introducing blockchain-based auditable logging \cite{YU2023100589}. 
These studies improve deployment feasibility, but they do not directly address real-time privacy budget allocation on terminal devices under dynamic multi-factor risk, nor do they establish a feedback-driven closed loop for continuous privacy–utility coordination in MECS.
A detailed comparison between ALPINE and representative prior studies is provided in the Appendix.

\begin{figure*}[t]
    \centering
    \includegraphics[width=\textwidth]{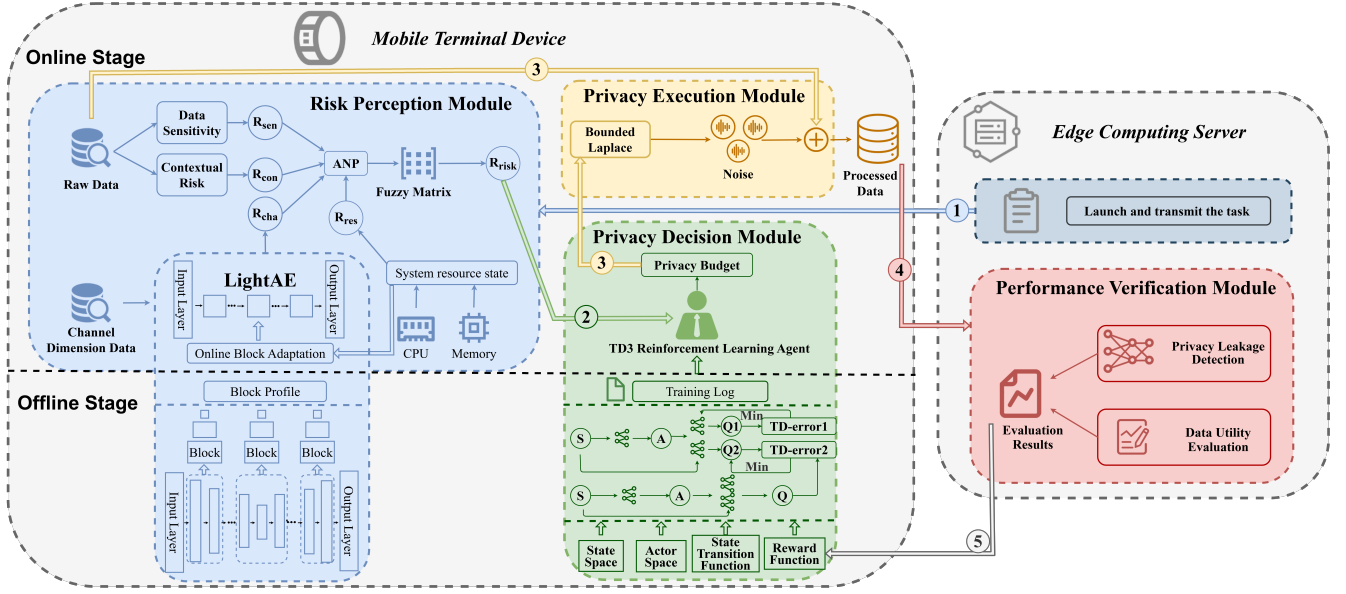}
    \caption
    {ALPINE is an adaptive lightweight framework for closed-loop privacy budget allocation in MECS. 
     A closed-loop control process: 
     (1) Server launches task;
     (2) Environmental risk score is forwarded to decision agent;
     (3) Noise is injected according to the decision; 
     (4) Processed data are transmitted to the server for validation; 
     (5) Validation results are fed back. }
    \label{fig:alpine-framework}
\end{figure*}
\section{Proposed Framework}
\subsection{Threat Model}
\subsubsection{Adversarial Roles and Capabilities}We consider two primary adversaries:
\textbf{External eavesdropper}, which monitors wireless channels, captures data transmitted from terminals to the edge server, and may perform signal sniffing, traffic analysis, and man-in-the-middle attacks.
\textbf{Honest-but-curious edge server}, executes the protocol faithfully, yet--out of commercial interest or curiosity--may analyze received data to infer sensitive information about individuals. We assume correct protocol execution; fully compromised terminals and collusion are outside our scope.

\subsubsection{Privacy Threats and Attack Vectors}We focus on the following privacy threats:
\textbf{Transmission-layer eavesdropping}. An adversary monitors wireless channels to capture data packets in transit. Weak signals and unstable links raise interception success.
\textbf{Data-level inference attacks}. The adversary exploits legitimately obtained data to infer sensitive information. These include:
Membership Inference Attack (MIA) \cite{7958568}: 
An adversary has partial background knowledge from public data and attempts to determine whether a queried record appeared in the training set. 
Property Inference Attack (PIA) \cite{10.1145/3243734.3243834}:
An adversary trains an auxiliary model on public data to infer sensitive properties from perturbed data.
Reconstruction Attack \cite{10.1145/2810103.2813677}:
An adversary exploits public data distributions and deep autoencoders to reconstruct perturbed data.
\textbf{Resource-oriented attacks}. By issuing bursty requests or malicious flooding, the adversary elevates the terminal’s compute load, potentially degrading or disabling privacy protection.

\subsubsection{Protection Objectives}
To counter transmission-layer eavesdropping, sufficient noise must be injected before data leaves the device.
To resist data-level inference, the protection strength must be aligned with semantic risk; highly sensitive data require stricter safeguards.
To mitigate resource-exhaustion, the privacy mechanism must be aware of resource risk and capable of graceful priority downgrading under tight budgets. Accordingly, we model privacy protection as dynamic privacy budget allocation driven by channel, semantic, contextual, and resource risk.

\subsection{Proposed Framework}

This study proposes ALPINE, a closed-loop adaptive privacy budget allocation framework for MECS. At its core is a feedback-controlled system with four modules—risk perception, privacy decision, privacy execution, and performance verification—for end-to-end adaptive privacy control. The overall framework is shown in Figure 2, and the workflow proceeds as follows.

First, the edge server generates sensing tasks and launches them to terminal devices. 
Upon receiving a task, the device activates the risk perception module that establishes a risk-evaluation mechanism across four dimensions: channel, semantic, contextual, and resource. 
Concretely, channel indicators are fed into a lightweight LightAE model trained with block-level adaptive scaling to produce a channel anomaly score $\mathrm{R}_{\text {cha}}$. 
In parallel, the device performs semantic-level analysis on the collected raw data to obtain data sensitivity $\mathrm{R}_{\text {sen}}$ and the contextual risk $\mathrm{R}_{\text {con}}$.
The device then incorporates real-time resource status (memory footprint and CPU utilization) to quantify a resource risk $\mathrm{R}_{\text {res}}$.
Finally, these risks are fused via an Analytic Network Process (ANP) -based fuzzy comprehensive evaluation, yielding an integrated environmental risk $\mathrm{R}_{\text {risk}}$.



In the privacy decision module, the system formulates privacy-budget allocation as a Markov Decision Process (MDP). 
A set of TD3 policies is pre-trained offline to learn mappings from environmental risk to privacy budgets under different operating regimes. 
During online inference, the terminal selects (and, if needed, switches) among the pre-trained policies and rapidly outputs an appropriate privacy budget based on the current risk state.
In the privacy execution module, the allocated budget drives the Bounded Laplace (BLP) mechanism that perturbs raw sensing data with calibrated noise.
The noised data are then transmitted over the communication link to the edge server.

The performance verification module runs on the edge server and evaluates the received data along two dimensions: privacy strength and data utility.
Privacy strength is assessed offline through canonical attack simulations (e.g., membership inference, property inference), while data utility is quantified via performance on downstream tasks.
The server converts these evaluations into feedback signals that are transmitted to the terminal for real-time policy switching and periodic drift detection, which triggers offline refinement and policy-set refresh when necessary. The four modules are detailed next.


\section{Proposed Technical Approach}

\subsection{Risk Perception Module}

\subsubsection{Channel Risk Modeling}We define the channel risk score $\mathrm{R}_{\text{cha}}$ to quantify the security and stability of wireless data transmission by integrating three indicators. 
Received Signal Strength Indicator measures the signal strength in receivers.
Link Quality reflects the stability and reliability of the communication channel. 
Delay Jitter measures via the round-trip time of ICMP packets.

To achieve efficient and accurate channel anomaly detection under resource constraints, we design a block-granularity scalable \emph{LightAE}. 
Motivated by dynamic, heterogeneous edge conditions, where compute and latency budgets fluctuate, a single fixed lightweight model cannot deliver an optimal accuracy–efficiency trade-off.
We adapt the idea of LightDNN \cite{9665270} and employ Autoencoder-based \cite{LI2023110176} block-level scaling: 
the network is partitioned into blocks with offline compressed descendants, and online we select the optimal combination under resource and latency constraints.

The architecture of LightAE is shown in Figure 3.
The method follows a two-stage pipeline: offline preparation and online optimization. 
Offline, we first train a full autoencoder and partition it into blocks, caching each block’s input/output tensors. 
For each block $i$, we generate a family of compressed descendant variants via structured pruning, and train each descendant to regress the intermediate representations of its corresponding original block (i.e., local knowledge distillation). We profile every variant to obtain a tuple
$\left(M_{i,j}, T_{i,j}, U_{i,j}\right)$,
where $M_{i,j}$ and $T_{i,j}$ denote memory/storage and latency costs, and $U_{i,j}$ denotes the reconstruction-error increase (i.e., utility degradation) relative to the full model.
During online inference, given current memory and latency budgets, a lightweight predictor estimates per-block costs and selects a block combination that minimizes the total reconstruction-error increase under the constraints, while changing only a small number of blocks to limit switching overhead (formulated as an integer linear programming problem). 
The resulting assembled detector outputs an anomaly score $\mathrm{s}(\mathrm{x})$ from real-time channel measurements, from which $\mathrm{R}_{\text{cha}}$ is derived.

\begin{figure}
  \centering
  \includegraphics[width=\linewidth]{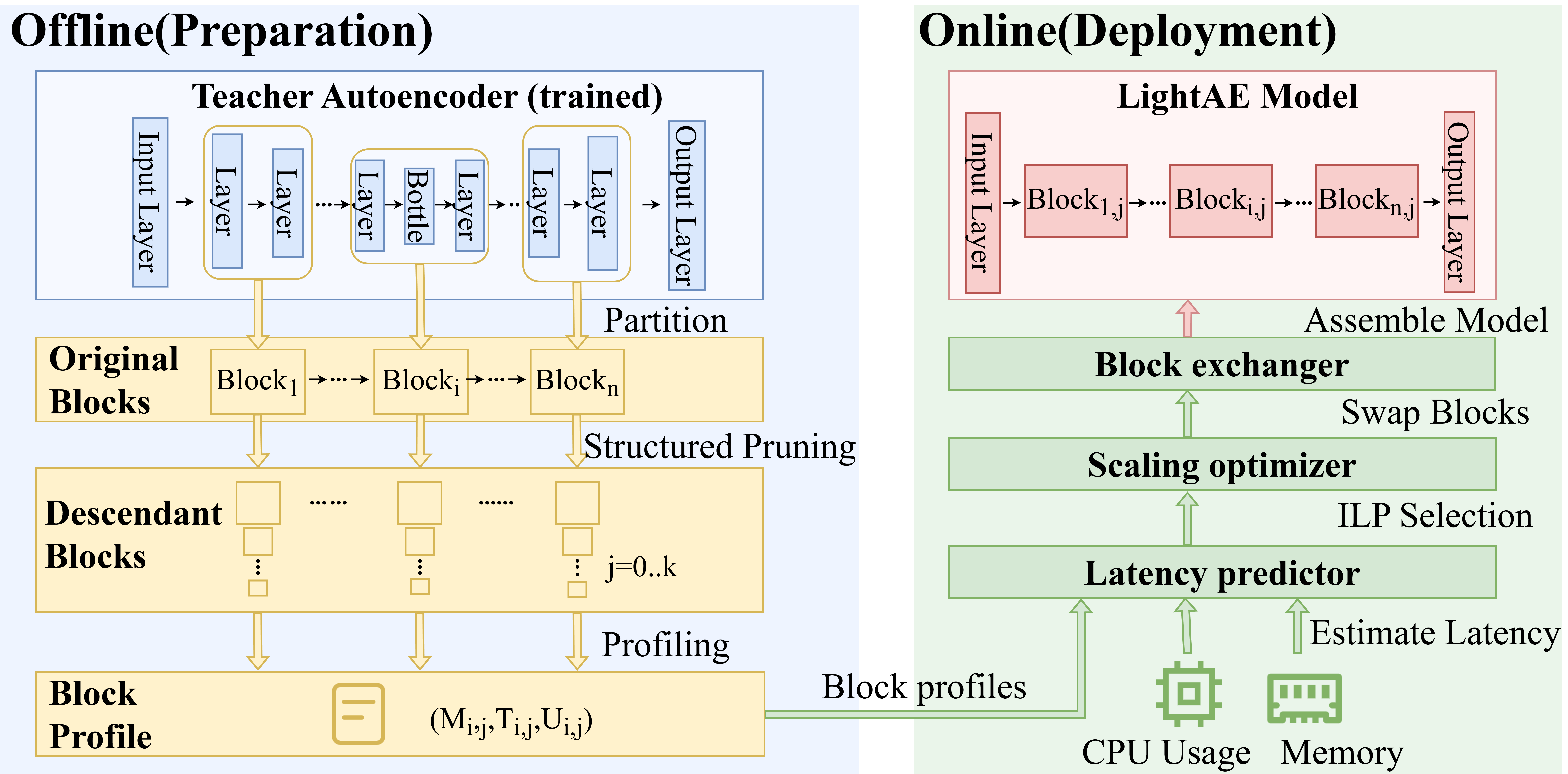}
  \caption{LightAE with block-granularity scaling. }
  \label{fig:lighae}
  \vspace{-6pt}
\end{figure}

\subsubsection{Semantic Risk Modeling}
Semantic risk captures privacy leakage caused by both a datum’s intrinsic sensitivity and its contextual associations; accordingly, it consists of data sensitivity and contextual risk.

Data sensitivity reflects the inherent sensitivity level of each field and is categorized by its data type. For example, location, health, and environmental data can be assigned sensitivity scores of 1.0, 0.8, and 0.3, respectively. Classification criteria can also draw from regulatory standards such as the General Data Protection Regulation (GDPR).
Many types of data are considered to have high sensitivity and should be treated accordingly in different application scenarios \cite{regulation2018general}.
Finally, we obtain the data-sensitivity risk $\mathrm{R}_{\text {sen}}$.

Contextual risk quantifies the entropy amplification effect that arises when a field co-occurs with other sensitive information in a specific context. 
Adversaries can exploit such contextual correlations to infer user privacy with greater accuracy \cite{chamikara2020efficient}. 
This risk is formally defined as follows:
\begin{equation}
\mathrm{R}_{\text {con }} =\frac{1}{\mathrm{n}} \sum_{\mathrm{i}=1}^{\mathrm{n}} \mathrm{I}\left(\text {Associated-field}_{\mathrm{i}}\right) \cdot \mathrm{H}\left(\mathrm{X}_{\mathrm{i}}\right),
\label{eq:isolation-d}
\end{equation}
$\text {Associated-field}_{\mathrm{i}}$ denotes the $\mathrm{i}$-th sensitive-associated field, and $\mathrm{X}_{\mathrm{i}}$ its corresponding random variable.
$\mathrm{I}(\cdot)$ is the sensitivity indicator function, and $\mathrm{H}(\cdot)$ the entropy quantifying uncertainty.

\subsubsection{Resource-usage Risk Modeling}
Terminal IoT devices have limited compute and storage; bursty requests or malicious processes can rapidly exhaust resources, causing latency spikes or denial of service.
Consequently, real-time monitoring of resource usage is critical for risk assessment and anomaly detection.
Methods for obtaining resource-usage data differ by device class \cite{ALWAISI2024101398}.
To quantify the impact of resource usage on risk, we adopt a joint metric of memory and CPU utilization.
$\mathrm{R}_{\text {res}}$ is computed as follows:
\begin{equation}
  \mathrm{R}_{\text {res}} =  \max \left(\frac{\text {MEM}_{\text{usa}}-\text {MEM}_{\text {nor}}}{\text {MEM}_{\max }-\text {MEM}_{\text {nor}}}, \frac{\text {CPU}_{\text{usa}}-\text {CPU}_{\text {nor }}}{\text {CPU}_{\max}-\text {CPU}_{\text {nor}}}\right),
\end{equation}
where ${\text {MEM}_{\text{usa}}}$ and ${\text {CPU}_{\text{usa}}}$ denote the real-time utilization, $\text {Mem}_{\text {nor}}$ and $\text {CPU}_{\text {nor}}$ are baseline averages under normal operating conditions, and $\text {Mem}_{\max }$ and $\text {CPU}_{\max }$ are the device’s physical or empirically determined upper bounds.
This design ensures a timely, conservative response to any bottleneck and prioritizing system stability and sustained privacy protection.

\subsubsection{Multi-dimensional Risk Perception Scoring}
We combine the ANP with fuzzy comprehensive evaluation. 
ANP is suited to complex systems in which criteria exhibit interdependence and feedback, allowing criteria to form a network structure \cite{Saaty2004}. 
Fuzzy comprehensive evaluation maps qualitative judgments into quantitative scores via membership functions and fuzzy rules \cite{chamikara2020efficient}.

First, we conduct ANP network analysis. 
We construct the network structure by accounting for the interdependence between any two risk dimensions.
Using the Saaty 1–9 scale for pairwise comparisons and experts provide judgments to form the pairwise comparison matrix. 
Applying the eigenvector method, we obtain the weights of the four dimensions: $\boldsymbol{\omega} = (\omega_{\mathrm{cha}},\, \omega_{\mathrm{sen}},\, \omega_{\mathrm{con}},\, \omega_{\mathrm{res}})$.

Next, we perform fuzzy comprehensive evaluation. 
We define an evaluation set and specify risk grades: $V=\left\{v_{1}, v_{2}, v_{3}\right\}$, along with numeric intervals. 
On this basis, membership functions are used to compute, for each dimension, the membership degree of a given risk score to each grade. 
By mapping the risk score to membership degrees via the membership function, we obtain a membership vector for that dimension.          
Stacking the membership vectors of all dimensions yields the fuzzy relation matrix $\mathrm{W}$:

\begin{equation}
    \mathrm{W}=\begin{bmatrix}
\mu_{c h a}^{1} & \mu_{cha}^{2} & \mu_{cha}^{3} \\
\mu_{s e n}^{1} & \mu_{s e n}^{2} & \mu_{s e n}^{3} \\
\mu_{c o n}^{1} & \mu_{c o n}^{2} & \mu_{c o n}^{3} \\
\mu_{r e s}^{1} & \mu_{r e s}^{2} & \mu_{r e s}^{3}
\end{bmatrix}.
\end{equation}
The matrix $\mathrm{W}$ reflects, for each risk dimension, memberships over the predefined risk grades. Multiplying the ANP weight vector by $\mathrm{W}$ yields the fuzzy synthesis: $B=\boldsymbol{\omega} \cdot \mathrm{W}=\left(b_{\text {1}}, b_{\text {2}}, b_{\text {3}}\right)$, where $b_{\text {i}}$ denotes the membership degree of the composite risk to three grades.

Finally, to convert the fuzzy result into a single scalar risk score, we apply weighted-average defuzzification:

\begin{equation}
    \mathrm{R}_{\text {risk }}=\frac{a_{\mathrm{1}} \cdot b_{\mathrm{1}}+a_{\mathrm{2}} \cdot  b_{\mathrm{2}}+a_{\mathrm{3}} \cdot b_{\mathrm{3}}}{b_{\mathrm{1}}+b_{\mathrm{2}}+b_{\mathrm{3}}},
\end{equation}
where $a_{\text {i}}$ denotes the corresponding grade, typically chosen as the centroid of each fuzzy set or set by expert knowledge. 
The resulting $\mathrm{R}_{\text {risk }}$ summarizes the overall system risk level.

\subsection{Privacy Decision Module}

We cast privacy-budget selection as an RL problem and train a TD3-based policy offline. 
During deployment, the terminal performs lightweight online inference to select a budget for the current risk state, enabling fast adaptation without online policy learning.

\subsubsection{MDP Modeling}
We formulate privacy budget allocation problem as a five-tuple $\mathrm{MDP}=(S, A, P, R, \gamma)$. The state is the continuous risk score $\mathrm{R}_{\text {risk }} \in [0,1]$. The action is the continuous privacy budget $\epsilon \in[\epsilon_{\min }, \epsilon_{\max }]$.
We use a smooth stochastic risk-dynamics transition $P\left(s_{t+1} \mid s_{t}, \epsilon_{t}\right)$ to capture temporal variability. 
The reward function jointly balances privacy-protection gain, data-utility loss and energy cost: 
\begin{equation}
R=\alpha \cdot \text {PrivacyGain}-\beta \cdot \text {UtilityLoss}-\lambda_{E} \cdot \text {EnergyCost}.
\label{eq:isolation-g}
\end{equation}

The privacy gain uses a logistic–power hybrid formulation.
In (6), $\kappa$ controls the steepness of the logistic curve, $s_{0}$ represents the predefined center and $\delta$ is the exponent-based budget penalty coefficient. 
The utility loss is explicitly linked to the expected distortion caused by the BLP, since the variance of the added noise scales as $1 / \epsilon^{2}$.
Therefore, we use a quadratic penalty, reflecting the statistical degradation of data utility as the privacy budget tightens.
$\rho$ is risk coupling coefficient and $\mathrm{g}_{0}$ is data sensitivity constant. 
Energy cost is measured with a power meter by integrating power over a time window.
$P$ denotes the instantaneous power and $\bar{E}_{t}$ denotes the average energy within the window.
Finally, the discount factor $\gamma$ computes the long-term cumulative reward.

\begin{equation}
\begin{aligned}
\text{PrivacyGain}&= \frac{1}{1+\exp\!\left[-\kappa\left(s_t-s_{0}\right)\right]}
\left(\frac{\varepsilon_{\max}-\varepsilon_t}{\varepsilon_{\max}-\varepsilon_{\min}}\right)^{\delta},\\
\text{UtilityLoss}&= (1-\rho\cdot s_t)\left(\frac{g_{0}}{\varepsilon_t}\right)^{2},\\
\text {EnergyCost}&=\bar{E}_{t} =\frac{1}{\Delta t} \int_{t}^{t+\Delta t} P(\tau) d \tau.
\end{aligned}
\label{eq:isolation-h}
\end{equation}

\subsubsection{TD3 algorithm}
To enable risk-adaptive allocation of the privacy budget, we employ the TD3 algorithm to build the policy agent. 
Given the current state $s$, the actor outputs a deterministic action $a=\mu(s \mid \theta^{\mu})$, which is mapped to the privacy budget for the current release.
TD3 belongs to the actor–critic family, comprising an actor network and two critic networks. 
The twin critics $Q_{1}(s, a\mid\theta^{Q_{1}})$ and $Q_{2}(s, a\mid\theta^{Q_{2}})$ estimate state-action values independently, and the minimum is used as the target Q-value, effectively suppressing overestimation bias \cite{pmlr-v80-fujimoto18a}.
The algorithm uses experience replay to decorrelate samples, and target networks with soft updates to stabilize training. During exploration, truncated Gaussian noise is injected into the action space to balance exploration and exploitation.
The TD3 agent learns an $\epsilon$-allocation policy over risk states to maximize the expected cumulative reward under privacy constraints.

\subsection{Privacy Execution Module}

Bounded Laplace (BLP) Mechanism. 
BLP guarantees that perturbed data fall within a prescribed interval $[l, u]$.
Given an input $x \in[l, u]$ and a scale parameter $b>0$, the probability density function (pdf) of BLP is defined as: 
\begin{equation}
    f_{w}\left(x^{*}\right)=\left\{\begin{array}{ll}
\frac{1}{C(x)} \frac{1}{2 b} \exp \left(-\frac{\left|x^{*}-x\right|}{b}\right), & x^{*} \in[l, u]\\
0, & x^{*} \notin[l, u]
\end{array}\right.,
\end{equation}
where $b=\Delta / \epsilon$ with $\Delta=u-l$ denoting the global sensitivity, $x^{*}$ is the noisy value, and $C(x)$ is a normalization constant ensuring that the pdf integrates to $1$ over $[l, u]$ \cite{7353177}.

To realize an efficient local-DP mechanism on terminal devices while respecting the natural bounds of sensor readings, we adopt BLP for noise injection. In the standard Laplace mechanism, the released value is generated as $x^{*}=x+\eta$ with $\eta \sim \operatorname{Lap}(0, b)$.
BLP re-normalizes the distribution over a prescribed interval, ensuring that the perturbed output always lies within a reasonable domain.
It preserves both validity and physical plausibility of released values while reducing abnormal boundary leakage. BLP is well suited to diverse sensing scenarios and sensor modalities.

\subsection{Performance Verification Module}


\subsubsection{Privacy-strength evaluation}
We construct three representative attackers---MIA, PIA and Reconstruction Attack, to validate the privacy protection.
These evaluations provide a direct indication of privacy-leakage risk and thus reflect privacy strength.

\subsubsection{Data-utility evaluation}
Utility evaluation measures how the perturbed data perform on specific downstream tasks. In this paper, we conduct binary-classification and regression experiments using public and historical datasets. The resulting downstream-task performance is used as the utility signal.

\subsubsection{Feedback mechanism}

We set target thresholds for privacy strength and data utility, and update the reward weights $(\alpha,\beta)$ in (5) as follows: 
\begin{equation}
\begin{aligned}
\alpha \leftarrow \Pi_{[\alpha_{\min},\alpha_{\max}]}\!\big(\alpha \cdot \exp(\eta_p e_p)\big)\\
\beta \leftarrow \Pi_{[\beta_{\min},\beta_{\max}]}\!\big(\beta \cdot \exp(\eta_u e_u)\big) ,
\end{aligned}
\end{equation}
where $\eta_p,\eta_u$ control the adaptation rate and $e_p,e_u$ measure deviations from the thresholds.
If privacy strength falls below its threshold, we increase $\alpha$; if utility falls below its threshold, we increase $\beta$. The updated weights are fed back to the terminal to select among offline-trained policies with different reward preferences, enabling fast online policy switching; persistent deviations can further trigger periodic offline retraining.

\section{Analysis and Evaluation}

\subsection{Privacy Analysis and Complexity Discussion}

\textbf{Theorem 1 }(Sequential Composition \cite{mcsherry2009privacy}).
If a sequence of local mechanisms $M_{1}, M_{2}, \ldots, M_{r}$ each satisfies $\epsilon_{i}-\text {LDP}$, then their composition $M$ satisfies $(\sum_{i} \epsilon_{i})-\text {LDP}$.

The theorem implies that privacy budgets can be allocated across mechanisms or features. 
In multi-sensor settings, per-sensor budgets $\epsilon_{i}$ can be assigned to temperature, humidity, illuminance, and current, enabling fine-grained privacy–utility trade-offs while respecting the overall local-DP constraint.

\textbf{Lemma 1}. In the proposed reward function, assuming a fixed energy window, there exists a unique global maximizer $\epsilon^{*}(s)$ at which the weighted marginal gains of privacy and utility are equal.

Lemma 1 further shows that our reward function satisfies the first-order Karush–Kuhn–Tucker optimality conditions for multi-objective optimization \cite{ghojogh2021kkt}, identifying the optimal point that balances privacy gain and utility loss.

In terms of cost and model complexity, ALPINE shifts the main computational burden to the offline stage. The online stage involves only forward passes of a few lightweight models, resulting in low and stable computational cost. Its storage demand is controllable and predictable: model parameters constitute a fixed post-deployment cost, while the lightweight design helps keep runtime memory usage low. Consequently, ALPINE is well suited for sustained operation on resource-constrained terminal devices.
The detailed proofs and analysis are provided in the Appendix.

\subsection{Experimental Analysis}

\subsubsection{Experimental Setup}
We construct a terminal–edge cooperative privacy-protection framework using Raspberry PI 5 as the terminal device and an edge server. 
Raspberry PI 5 has a Broadcom BCM2712 CPU (Cortex-A76, 2.4 GHz), 8GB LPDDR4X RAM and 32GB MicroSD storage. 
The edge server uses an Intel Core i9-14900K CPU (6 GHz), 128GB RAM and 2GiB swap space. The software stack includes Python and PyTorch, and communication is implemented via MQTT.

We use three datasets for channel-anomaly detection and three datasets for downstream performance evaluation. 
For channel dimension, we construct two perturbed and anomaly-injected channel datasets. The first dataset (\textbf{FD}) is collected from a Raspberry PI terminal and contains 24 hours of continuous network monitoring. The second dataset (SD) is collected from a laptop and records 40 hours of network activity. 
The test sets contain four types of simulated anomalies: physical-layer signal anomalies, network-layer transmission anomalies, hardware failures and adversarial attacks. 
In addition, we use the \textbf{public KDD CUP HTTP dataset} for generalization experiments, creating a low-dimensional feature subset to assess the model’s anomaly-detection performance \cite{protic2018review}.

For downstream tasks, we select three real-world datasets from IoT sensing, smart-home and healthcare domains. 
\textbf{Intel Berkeley Research Lab Sensor Data} is used for a binary-classification task with multi-feature inputs (temperature, humidity, light and voltage) to evaluate data utility. 
\textbf{UK-DALE dataset} is used for regression and classification tasks in non-intrusive load monitoring. 
\textbf{Diabetes 130-US Hospitals Dataset} is used for readmission prediction and readmission-window analysis, and is formulated here as a binary-classification task to evaluate data utility \cite{bhardwaj2024diabetic,liu2024comparison}.

\begin{table}[h]
  \centering
  \begin{threeparttable}
  \caption{Online Scaling of LightAE}
  \label{tab:ablation}

  \begin{tabular}{ccc}
    \hline
    (Latency , Resource) (\%) & F1 (\%) & Memory (MB) \\
    \hline
    (0, 0)    & 96.28 & 1.98 \\
    (20, 20)  & 95.96 & 1.66 \\
    (50, 50)  & 95.79 & 1.00 \\
    (50, 80)  & 95.42 & 0.82 \\
    (80, 50)  & 95.42 & 0.81 \\
    \hline
  \end{tabular}
  \vspace{-6pt}
  \end{threeparttable}
\end{table}

\subsubsection{Anomaly Detection Performance Evaluation}
We evaluate LightAE under varying network and resource conditions.

\begin{table*}[t]
  \centering
  \caption{Model Performance Comparison across Three Anomaly-Detection Datasets}
  \label{tab:fullwidth-benchmark}
  \begin{adjustbox}{max width=\textwidth}
  \begin{threeparttable}
  \begin{tabular}{l
                  ccccc
                  cccccc
                  ccccc}
    \toprule 
    & \multicolumn{5}{c}{FD Dataset}
    & \multicolumn{5}{c}{SD Dataset}
    & \multicolumn{5}{c}{HTTP Dataset} \\
    \cmidrule(lr){2-6}\cmidrule(lr){7-11}\cmidrule(lr){12-16}
    Model &
Prec. & Rec. & F1 & Mem. & Time &
Prec. & Rec. & F1 & Mem. & Time &
Prec. & Rec. & F1 & Mem. & Time \\
    \midrule
    IsolationForest & 77.97 & 76.10 & 77.03 & 1.10 & \underline{0.18} &
                      74.18 & 76.51 & 75.32 & 1.73 & \underline{0.39} &
                      63.90 & 91.03 & 75.09 & 1.51 & \textbf{1.12} \\
    One-Class~SVM   & 93.05 & 94.27 & 93.66 & \textbf{0.02} & \textbf{0.12} &
                      86.73 & 93.59 & 90.09 & \textbf{0.01} & \textbf{0.04} &
                      79.82 & 91.69 & 85.34 & \textbf{0.02} & \underline{2.83} \\
    LSTM            & 93.02 & 84.50 & 91.59 & \underline{0.05} & 37.49 &
                      91.09 & 93.72 & 92.38 & \underline{0.05} & 78.69 &
                      87.68 & 91.65 & 89.62 & \underline{0.05} & 659.60 \\     
    LSTM-NDT        & 97.36 & 94.20 & 95.75 & 0.25 & 68.40 &
                      95.66 & 92.51 & 94.06 & 0.26 & 250.62 &
                      86.82 & 91.65 & 89.17 & 0.26 & 2245.70 \\ 
    OmniAnomaly     & \textbf{99.98} & \textbf{96.50} & \textbf{98.20} & 0.35 & 135.82 &
                      96.36 & \textbf{95.80} & \textbf{96.08} & 0.35 & 268.40 &
                      87.68 & 91.65 & 89.62 & 0.35 & 2661.34 \\ 
    iTransformer     & 96.81 & 94.19 & 95.48 & 0.11 & 196.86 &
                      84.91 & 91.54 & 88.10 & 0.11 & 363.66 &
                      89.89 & 91.57 & 90.72 & 0.11 & 4228.44 \\
    ModernTCN        & 80.59 & 93.64 & 86.63 & 0.42 & 1052.48 &
                      77.99 & 88.77 & 83.03 & 0.42 & 1891.21 &
                      87.67 & 91.57 & 89.58 & 0.45 & 17864.40 \\
    Autoencoder      & \underline{98.45} & 93.73& \underline{95.96} & 1.98 & 142.55 &
                      \textbf{98.77} & \underline{93.42} & \underline{96.02} & 1.98 & 252.01 &
                      \textbf{90.04} & \underline{91.66} & \textbf{90.84} & 1.98 & 2018.16 \\     
    Autoencoder+Pruning        & 95.38 & 94.20 & 94.79 & 1.28 & 128.42 &
                      97.64 & 93.23 & 95.38 & 1.02 & 232.43 &
                      87.76 & \textbf{91.73} & 89.70 & 1.20 & 1937.82 \\ 
    Autoencoder+KD    & 96.44 & 93.20 & 94.79 & 0.52 & 166.24 &
                      96.07 & 93.10 & 94.56 & 0.56 & 277.66 &
                      87.74 & 91.63 & 89.63 & 0.55 & 2226.67 \\
    Autoencoder+CGNet     & 95.38 & \underline{94.28} & 94.83 & 0.64 & 140.33 &
                      98.00 & 92.98 & 95.43 & 0.62 & 245.93 &
                      \underline{89.88} & 89.65 & 89.76 & 0.66 & 1991.58 \\                      
                      
    LightAE         & 98.30 & 93.50 & 95.70 & 1.18 & 75.42 &
                      \underline{98.63} & 93.03 & 95.57 & 1.35 & 130.84 &
                      88.59 & 91.57 & \underline{90.05} & 1.37 & 560.32  \\

    \bottomrule
  \end{tabular}
\begin{tablenotes}
\item[] The \textbf{bold} indicates the best result and the \underline{underlining} denotes the second best.
\end{tablenotes}
\vspace{-6pt}
\end{threeparttable}
  \end{adjustbox}

\end{table*}

LightAE builds on a block-partitioned autoencoder and adapts online by selecting block variants to meet latency and memory constraints.
Table \ref{tab:ablation} reports model accuracy and size under different latency-convergence and resource-constraint percentages. Under light constraints, performance remains close to the baseline autoencoder; under tighter constraints, the controller selects lighter block variants. This accuracy change arises because stricter constraints force the system to switch to lighter descendant blocks. These lighter blocks have fewer parameters and simpler architectures, which inherently limits their feature extraction and reconstruction capacity, leading to a slight drop in detection performance. These results indicate that LightAE’s online scaling enables efficient adaptation across different constraint targets. 

In Table \ref{tab:fullwidth-benchmark}, we use comparable parameter budgets and training epochs across models and report averages over five runs.
The table reports precision (Prec.), recall (Rec.), F1-score (F1), memory usage in \textbf{megabytes (MB)} (Mem.), and training time in \textbf{seconds (s)} (Time).
Compared with traditional ML baselines (One-Class SVM \cite{scholkopf2001estimating}, Isolation Forest \cite{liu2008isolation}), LightAE achieves higher accuracy and stronger anomaly-detection sensitivity. 
Against deep time-series models (LSTM \cite{malhotra2015long}, LSTM-NDT \cite{hundman2018detecting}, OmniAnomaly \cite{su2019robust}, iTransformer \cite{liu2023itransformer}, ModernTCN \cite{luo2024moderntcn}), LightAE matches the top metrics without explicit sequence modeling or complex post-processing, while incurring markedly lower resource cost. 
Relative to the base autoencoder and its lightweight variants (pruning \cite{filters2016pruning}, knowledge distillation \cite{tian2019contrastive}, CGNet \cite{hua2019channel}), LightAE shows no substantial accuracy drop from the baseline and offers a superior accuracy–memory trade-off than pruning-only or KD-only versions. In summary, LightAE achieves a favorable balance among accuracy, stability, and resource efficiency.

\subsubsection{Effectiveness of the Dynamic Privacy Strategy}We compare three deep RL algorithms TD3, DDPG \cite{lillicrap2015continuous}, and Soft Actor-Critic (SAC) \cite{haarnoja2018sof}. 
The MDP configuration is held fixed across methods, and results are averaged over multiple runs.
\begin{figure}[h]
  \centering
  \begin{subcaptionblock}{0.48\linewidth}
    \centering
    \includegraphics[width=\linewidth]{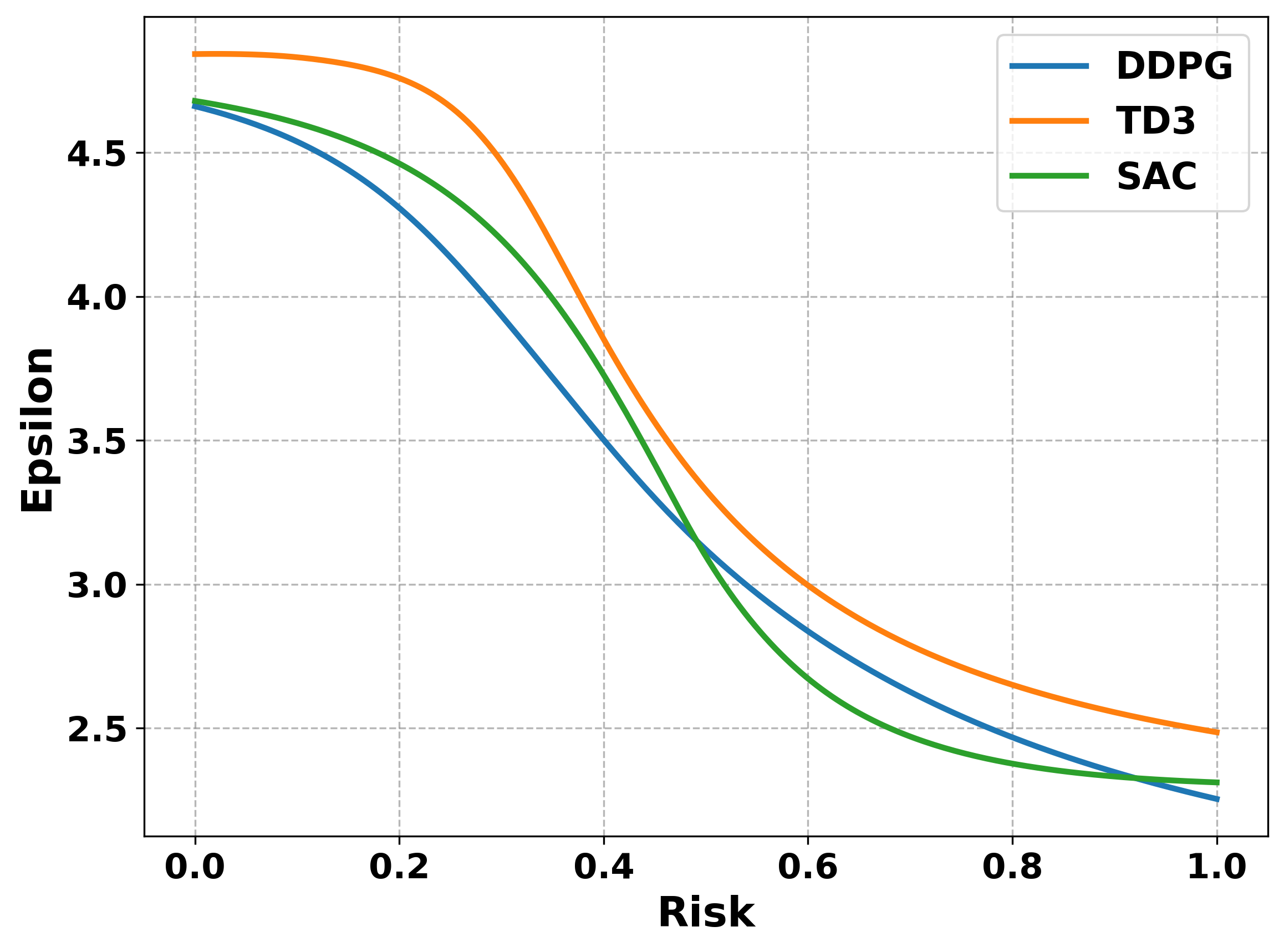}
  \end{subcaptionblock}\hfill
  \begin{subcaptionblock}{0.48\linewidth}
    \centering
    \includegraphics[width=\linewidth]{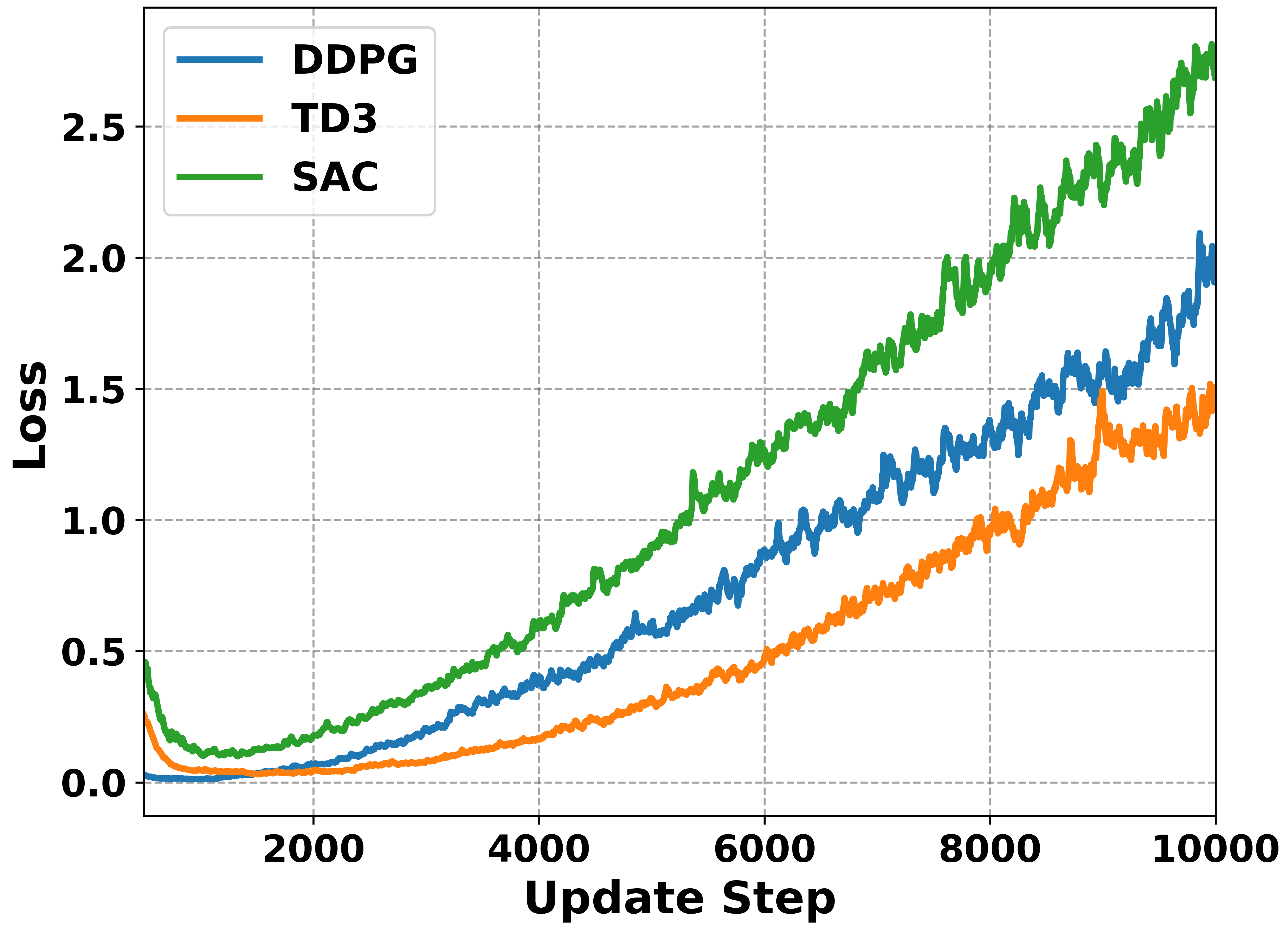}
  \end{subcaptionblock}
  \caption{The comparison of three RL models.}
  \label{fig:rl}
\end{figure}

In Figure 4, DDPG exhibits an approximately linear downward trend. SAC changes too abruptly in mid-risk regions, yielding a steeper decision boundary. By contrast, TD3 adjusts the policy more smoothly across risk levels, remaining sensitive to risk changes while maintaining better stability. Regarding loss convergence, DDPG converges steadily; SAC converges faster initially but shows larger later-stage oscillations. 
TD3 maintains the lowest loss trajectory with the smallest oscillations. Quantitatively, TD3 achieves the shortest average training time (34.5 s), compared with DDPG (35.4 s) and SAC (63.7 s), and also yields a higher mean reward, indicating more consistent privacy-budget policies across runs.

\begin{figure}[h]
  \centering
  \begin{subcaptionblock}{0.48\linewidth}
    \centering
    \includegraphics[width=\linewidth]{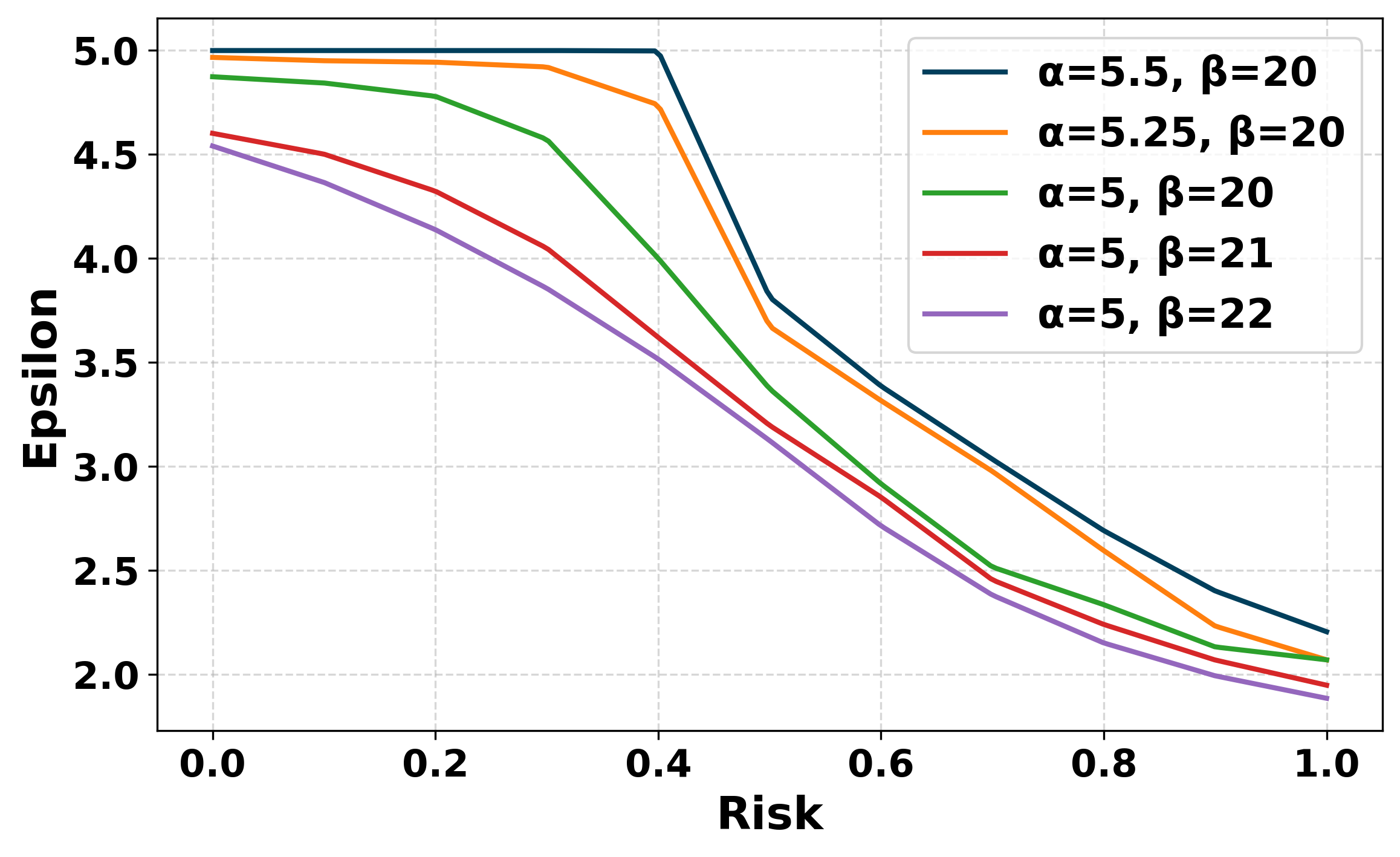}
  \end{subcaptionblock}\hfill
  \begin{subcaptionblock}{0.48\linewidth}
    \centering
    \includegraphics[width=\linewidth]{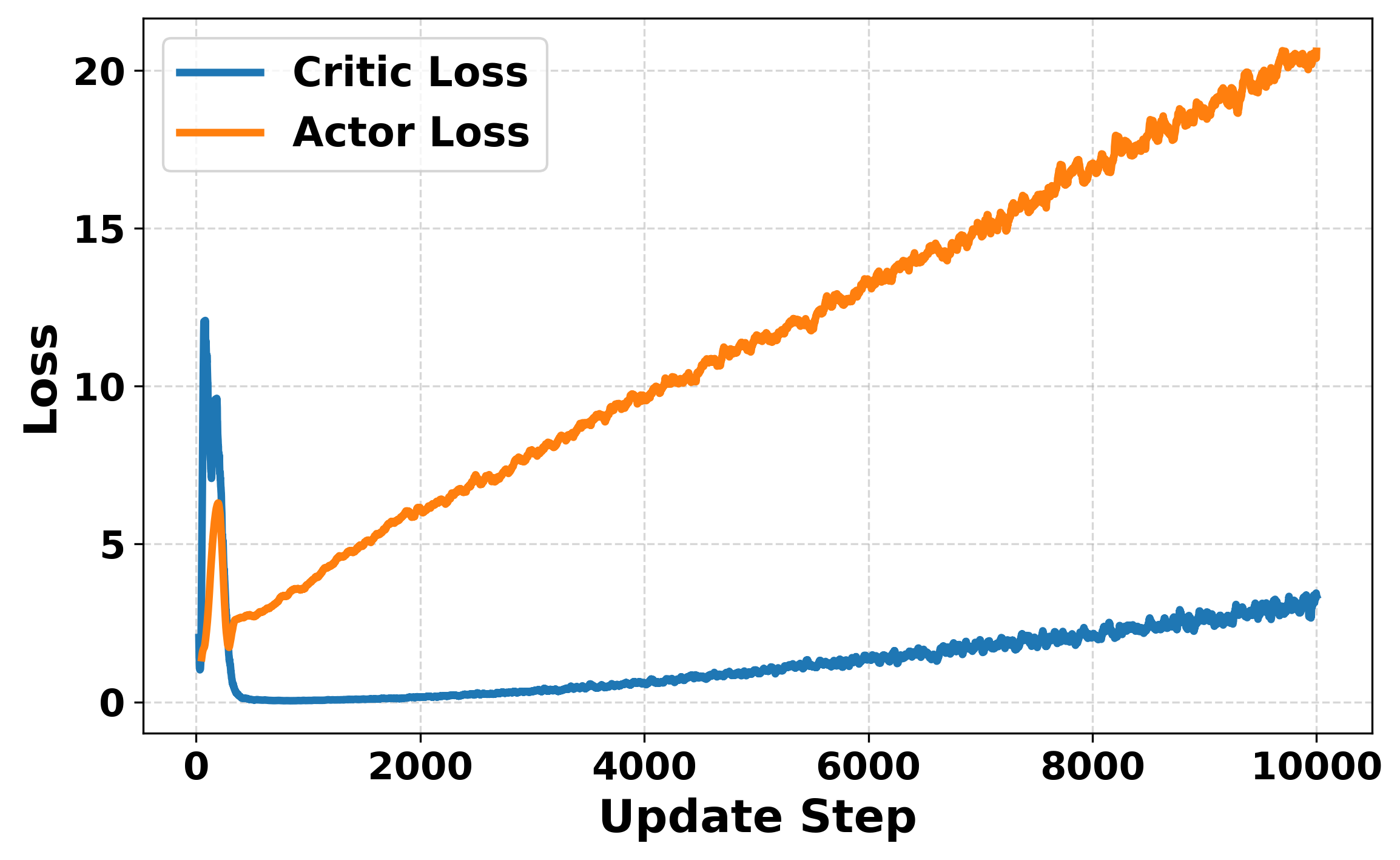}
  \end{subcaptionblock}
  \caption{The performance of TD3 model.}
  \label{fig:rl2}
\end{figure}
We further analyze TD3 by offline training a policy set with different reward weights $(\alpha,\beta)$ to validate the feedback mechanism. 
Larger $\alpha$ yields tighter budgets (more noise), while larger $\beta$ relaxes budgets (less noise) (Figure 5), thereby forming a controllable policy family for online switching.
This shows that tuning $\alpha$ and $\beta$  effectively steers the privacy-budget allocation, enabling a flexible trade-off between privacy protection and data utility.
The right panel shows that TD3 achieves favorable Actor–Critic loss convergence.
Although the critics fluctuate at the beginning, they quickly converge to near zero within the first 1{,}000 steps, indicating stabilized Q-value estimates.
Meanwhile, the actor loss increases gradually but remains overall steady, reflecting continued refinement of the policy’s action outputs during training.

\subsubsection{Performance Verification Analysis}
We conduct systematic evaluations on real-world datasets to validate the balance between data utility and privacy strength.

\begin{figure}[h]
  \centering
  \begin{subcaptionblock}{0.48\linewidth}
    \centering
    \includegraphics[width=\linewidth]{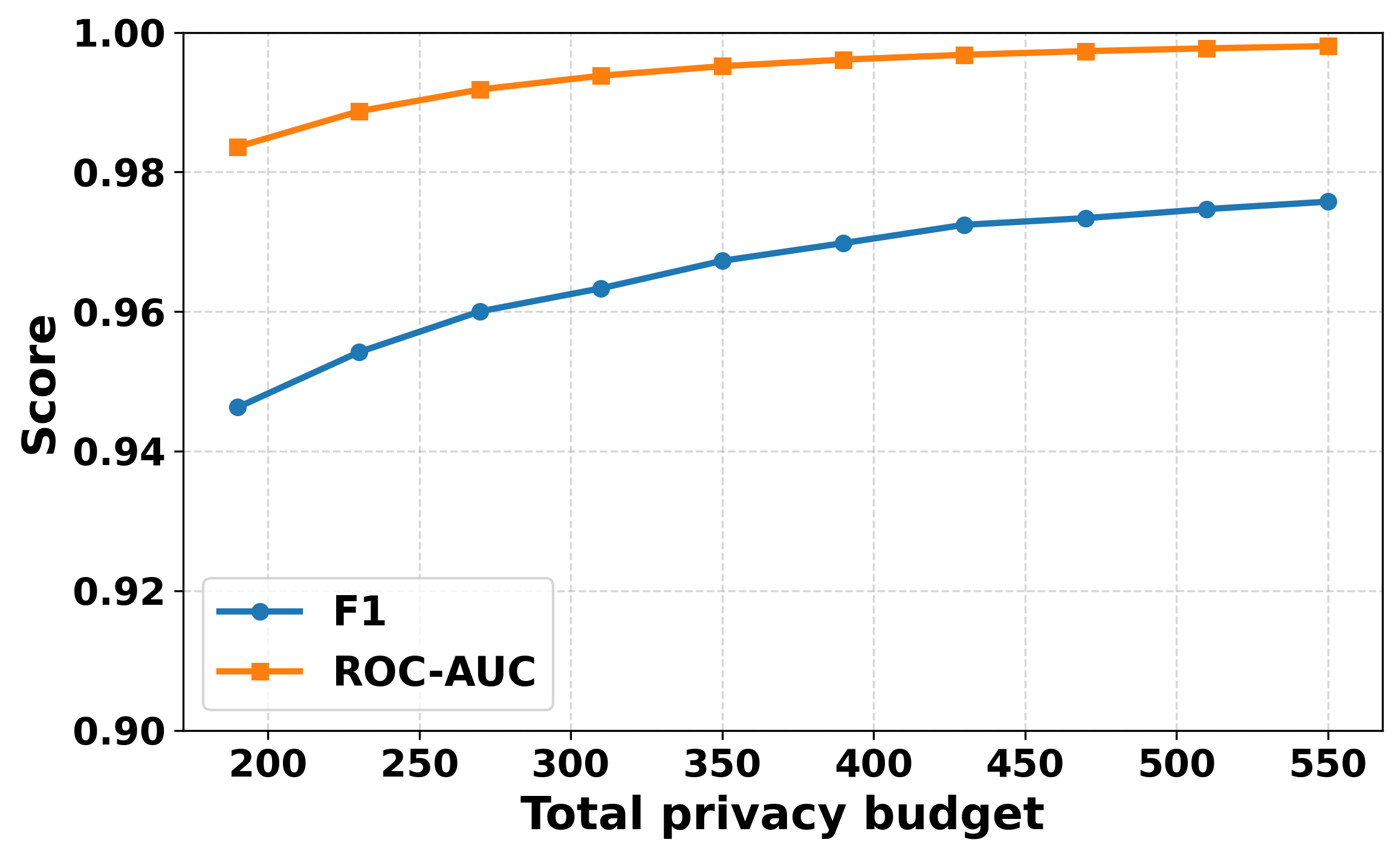}
  \end{subcaptionblock}\hfill
  \begin{subcaptionblock}{0.48\linewidth}
    \centering
    \includegraphics[width=\linewidth]{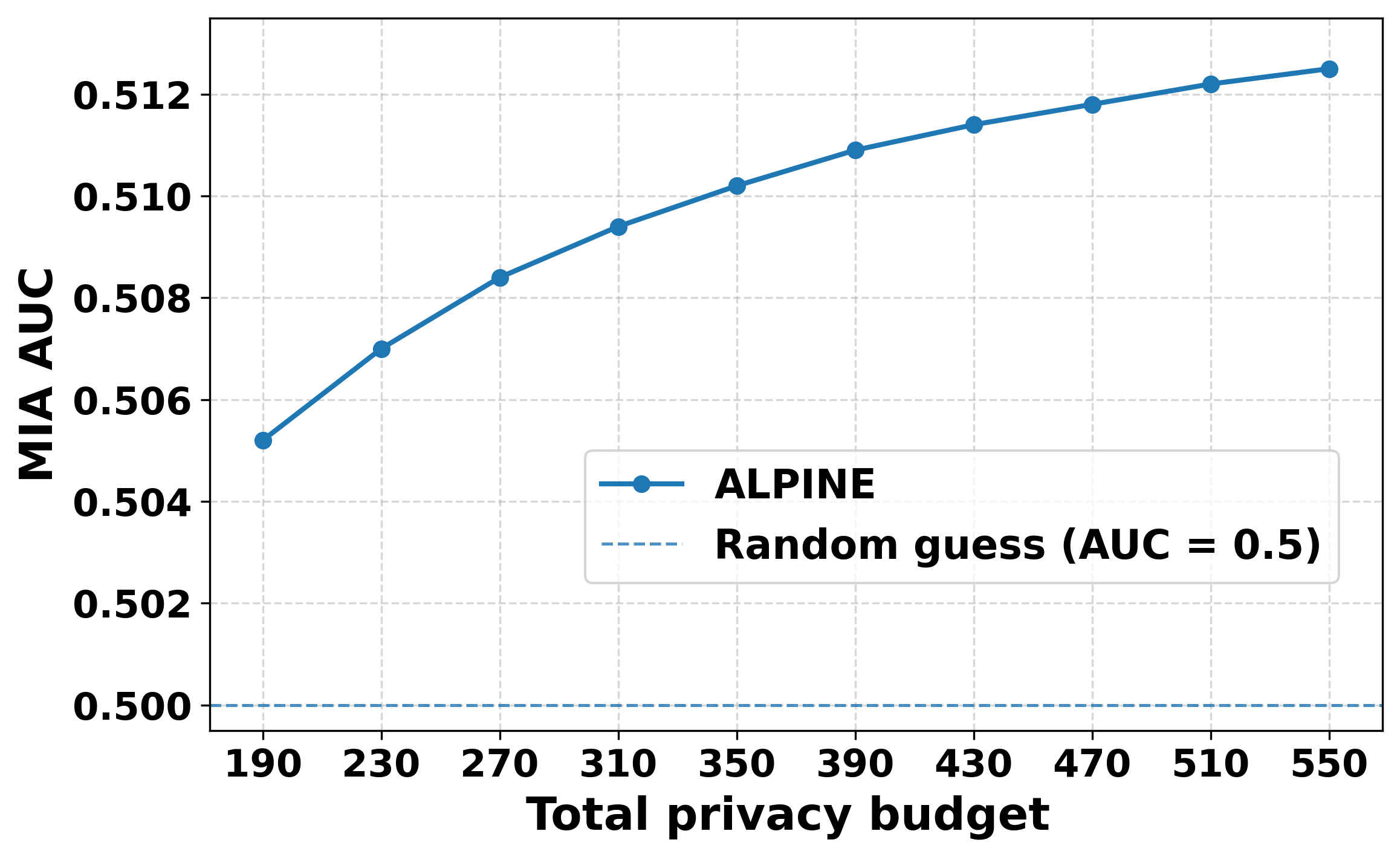}
  \end{subcaptionblock}
  \caption{Evaluation in Intel Berkeley Research Lab Sensor Data.}
  \label{fig:d2}
\end{figure}

For the Intel Berkeley Research Lab Sensor Data, we use Light as the primary prediction target and construct a binary-classification task from its binarized labels, with Temperature, Humidity, and Voltage as auxiliary features. Under varying privacy budgets, we assess both data utility and resilience to MIA.
We report F1-score and ROC-AUC as the utility metrics. 
Figure 6 shows that model performance improves as the privacy budget increases, whereas heavy noise degrades both metrics.
To evaluate privacy robustness, we construct an MIA model based on prediction confidence and measure attack effectiveness using AUC. The right panel of Figure 6 shows that the attack AUC remains close to random guessing (AUC $=0.5$) across the entire budget range, although it increases slightly as the privacy budget grows. 
This result indicates that ALPINE maintains effective resistance to MIA while gradually improving task utility under larger privacy budgets.
\begin{figure}[h]
  \centering
  \begin{subcaptionblock}{0.48\linewidth}
    \centering
    \includegraphics[width=\linewidth]{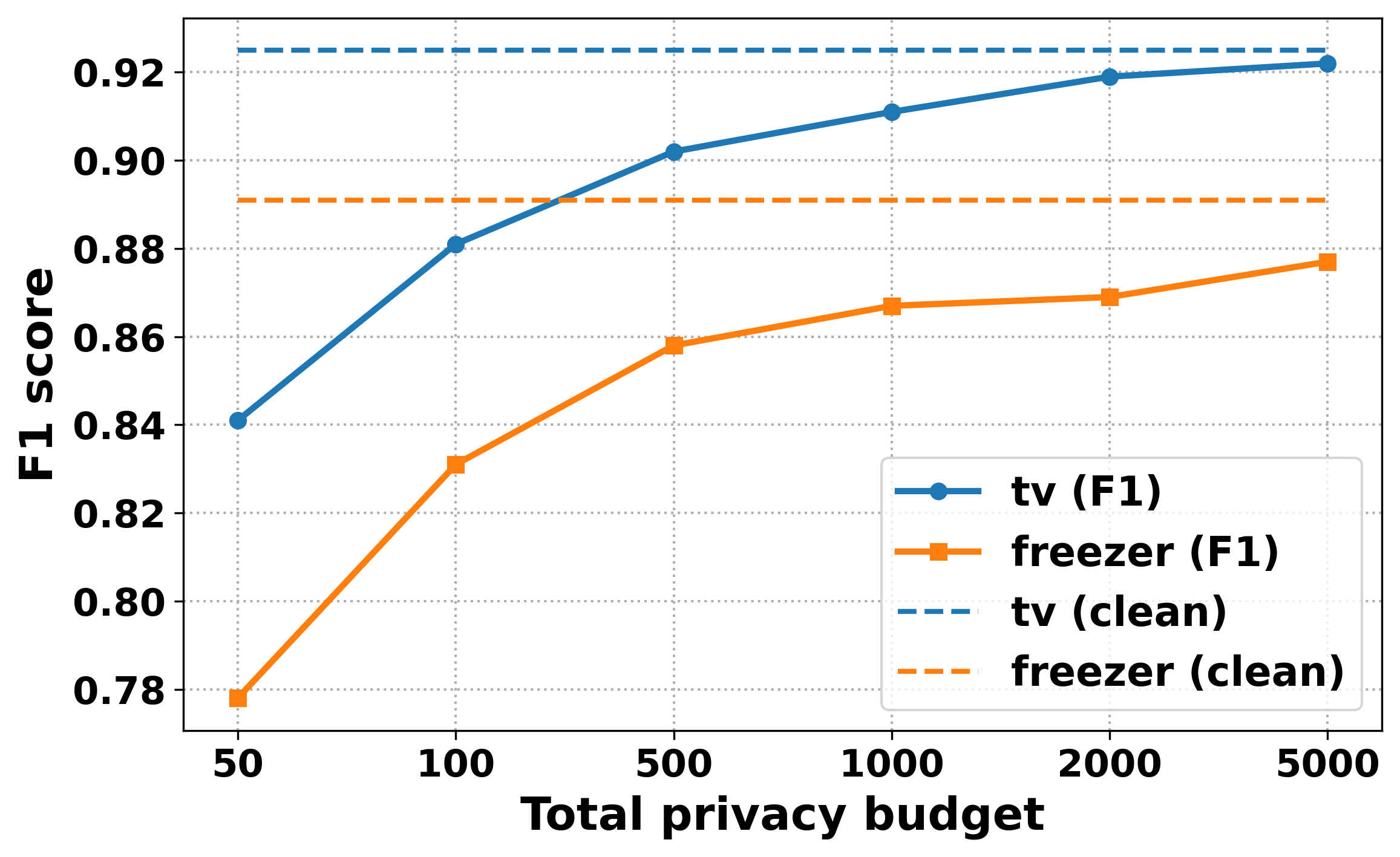}
  \end{subcaptionblock}\hfill
  \begin{subcaptionblock}{0.48\linewidth}
    \centering
    \includegraphics[width=\linewidth]{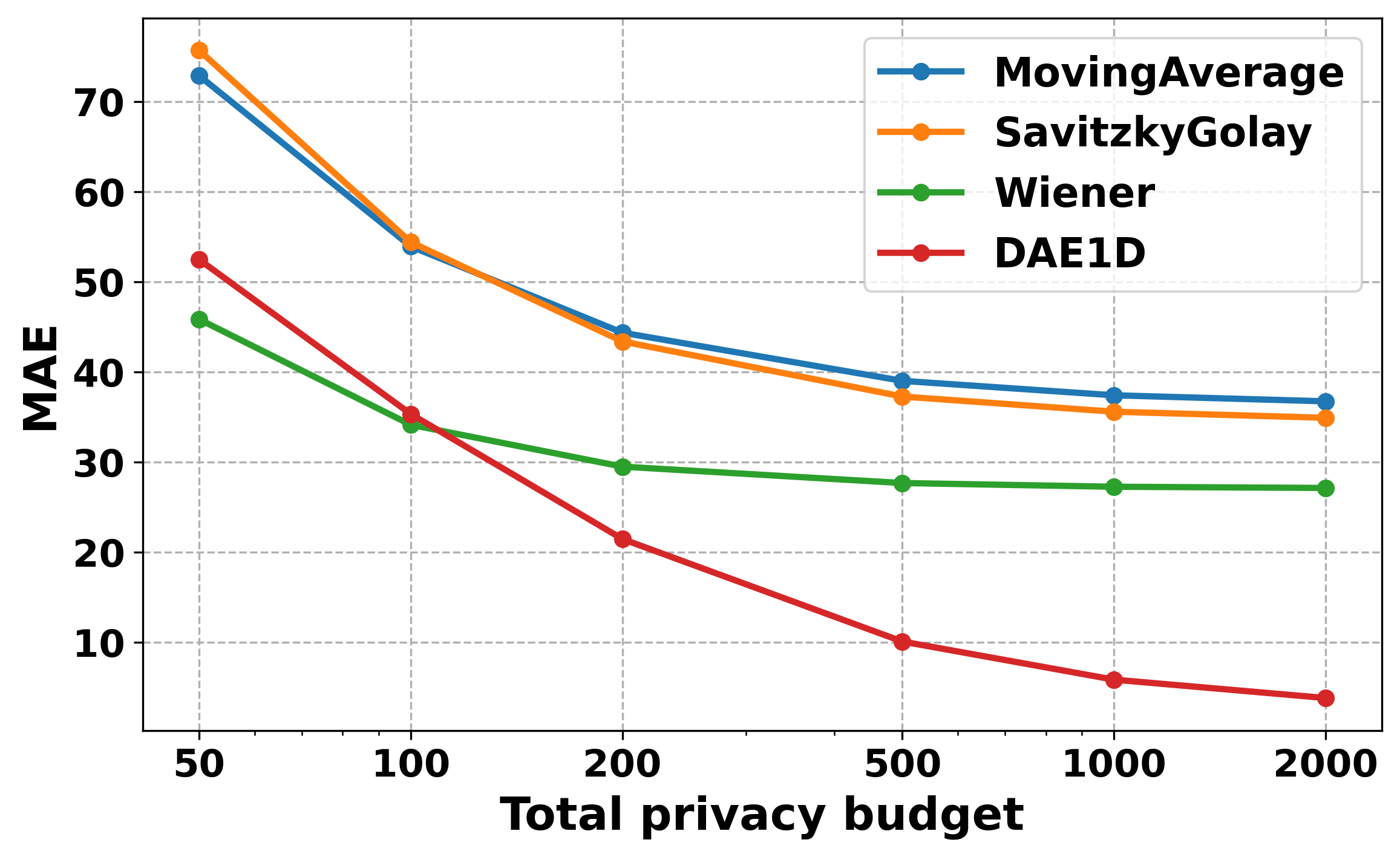}
  \end{subcaptionblock}
  \caption{Evaluation in UK-DALE.}
  \label{fig:d3}
\end{figure}

For UK-DALE, the dataset provides minute-level measurements of whole-home (aggregate) power and multiple appliance loads, with wide dynamic ranges and abrupt power variations for some devices. We assess the privacy–utility trade-off of BLP in a NILM setting and apply BLP only to the aggregate consumption stream. 

For NILM, we evaluate classification performance under varying privacy budgets by predicting the on/off states of two representative devices, television and freezer, using F1 as the metric. In Figure 7, as the privacy budget increases, the model approaches its performance at a clean level, indicating that it captures load characteristics more effectively.
Meanwhile, in reconstruction attacks, we evaluate four post-hoc denoising strategies: moving average, Savitzky–Golay smoothing\cite{schmid2022and}, Wiener filtering and a 1-D deep denoising autoencoder \cite{ahmed2019mitigating}. 
The mean absolute error (MAE) versus privacy budget $\epsilon$ curves show  BLP markedly strengthens resistance to reconstruction under small $\epsilon$.
\begin{figure}[h]
  \centering
  \begin{subcaptionblock}{0.48\linewidth}
    \centering
    \includegraphics[width=\linewidth]{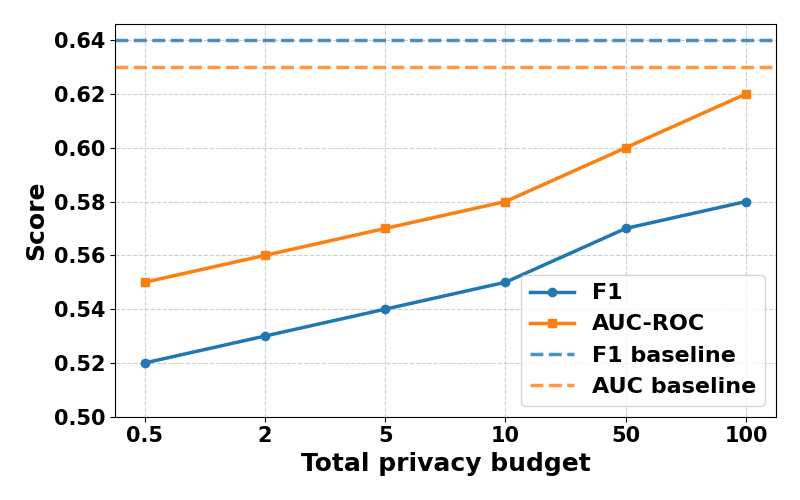}
  \end{subcaptionblock}\hfill
  \begin{subcaptionblock}{0.48\linewidth}
    \centering
    \includegraphics[width=\linewidth]{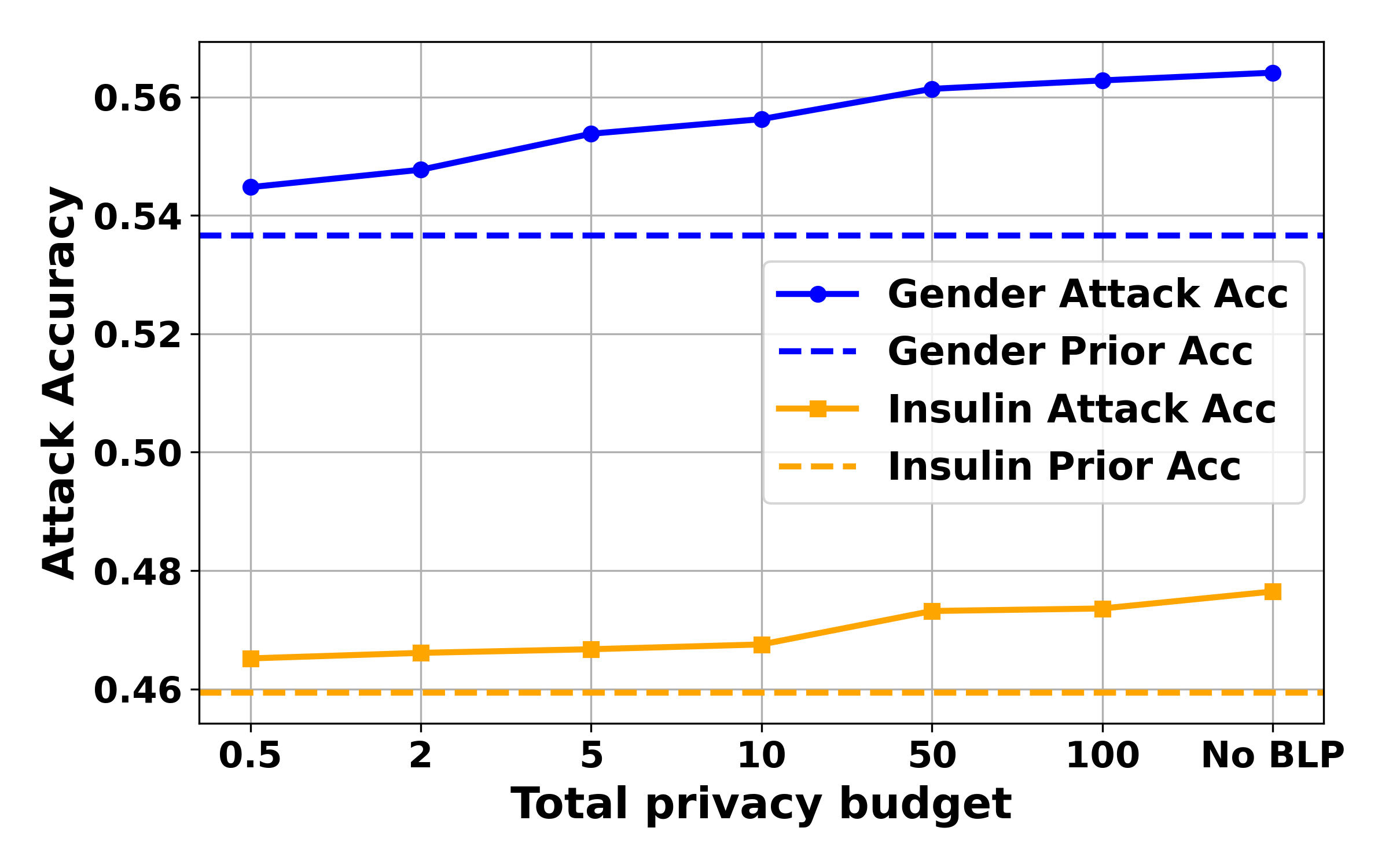}
  \end{subcaptionblock}
  \caption{Evaluation in Diabetes 130-US Hospitals dataset.}
  \label{fig:d4}
\end{figure}

For the Diabetes 130-US Hospitals dataset, we perturb eight continuous features with BLP under varying $\epsilon$ and train an XGBoost model for readmission prediction. To mitigate class imbalance, we apply SMOTE during training and report F1-score and AUC-ROC. 
Figure 8 shows that small $\epsilon$ noticeably degrades performance, while larger $\epsilon$ gradually recovers toward the no-noise baseline.
We also evaluate property-inference robustness: the target model is trained without sensitive attributes, and an adversary fits a logistic regressor using predicted probabilities and public features, using Prior ACC (marginal prevalence) as the reference.
Small $\epsilon$ keeps outcomes near the prior, whereas larger $\epsilon$ weakens protection.

Overall, across diverse scenarios, the BLP mechanism exhibits a consistent privacy–utility trade-off.
Under small privacy budgets, noise injection strengthens defenses against both membership inference and property inference, albeit with some loss of task utility. 
Under large budgets, predictive performance essentially returns to the noise-free baseline, while privacy protection progressively weakens. 
Meanwhile, different task types and feature distributions exhibit varying sensitivities to the privacy budget. 
In practical deployments, privacy-budget selection should remain context-aware and aligned with application scenarios and task requirements, with timely feedback provided when available.

\begin{figure*}[htbp]
  \centering
  \begin{minipage}{0.33\textwidth}
    \centering
    \includegraphics[width=\textwidth]{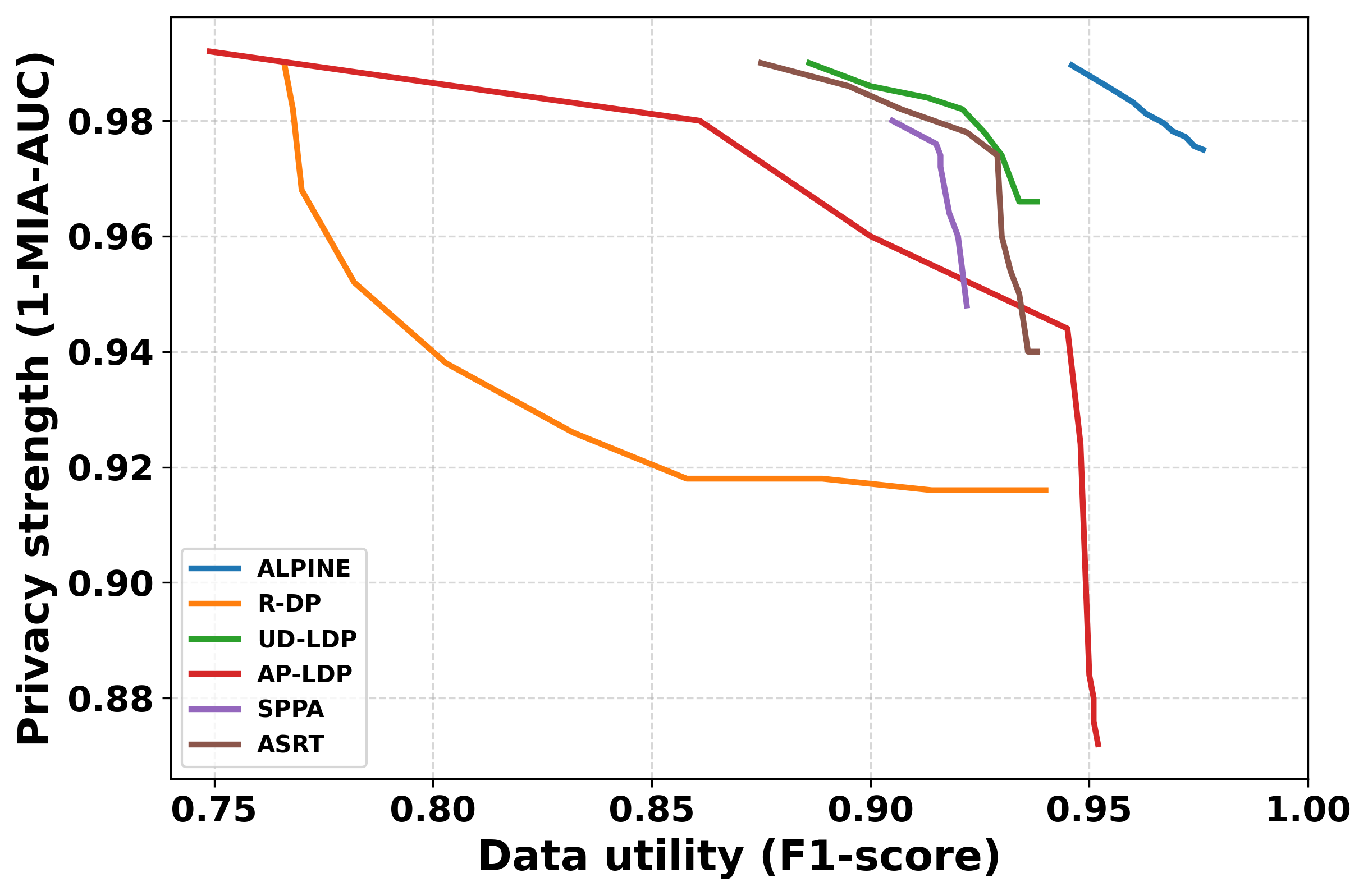} 
    \caption{Trade-off on Intel Dataset.}
    \label{fig:intel_privacy_utility}
  \end{minipage}%
  \begin{minipage}{0.33\textwidth}
    \centering
    \includegraphics[width=\textwidth]{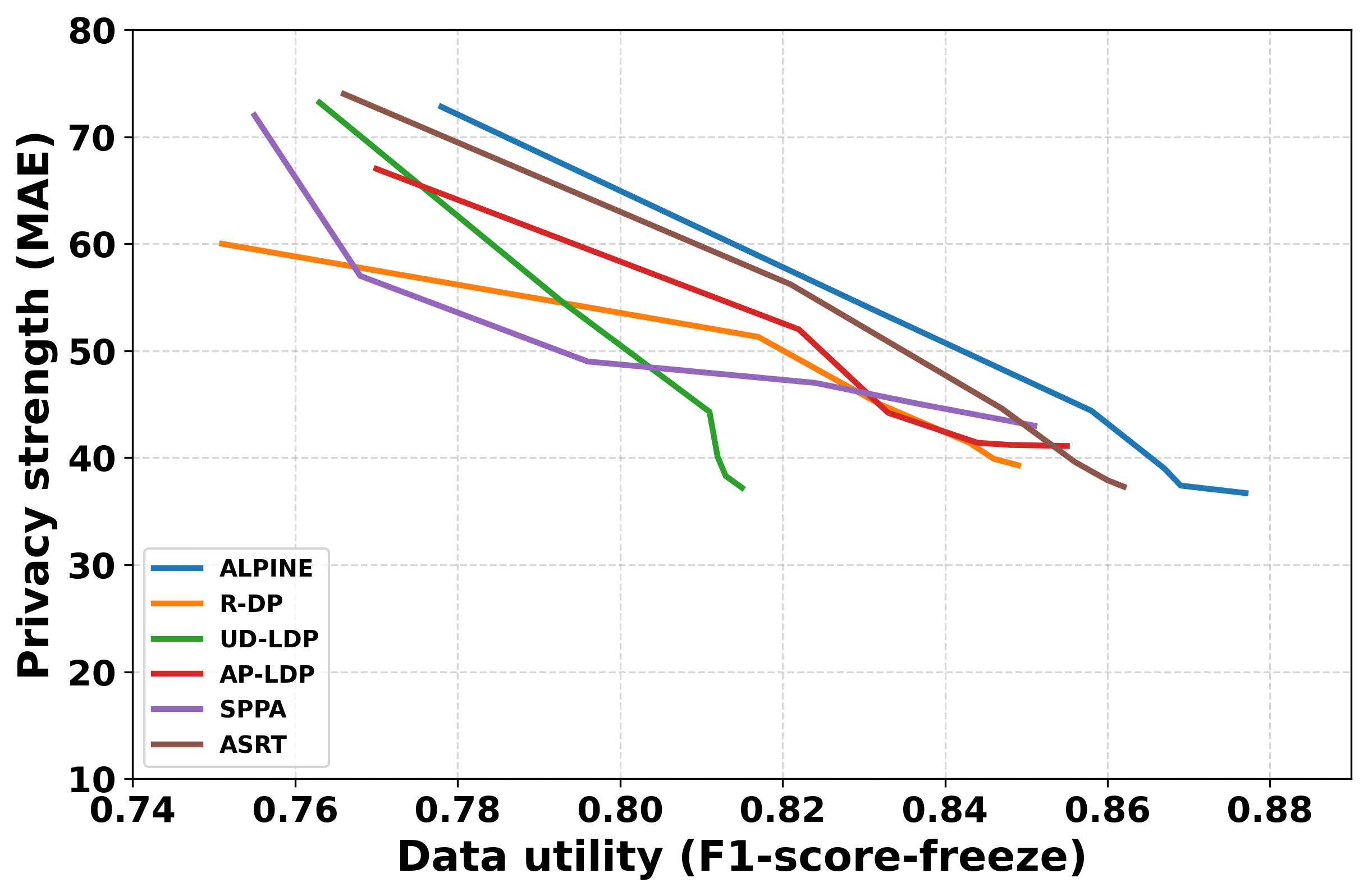} 
    \caption{Trade-off on UK-DALE Dataset.}
    \label{fig:ukdale_privacy_utility}
  \end{minipage}%
  \begin{minipage}{0.33\textwidth}
    \centering
    \includegraphics[width=\textwidth]{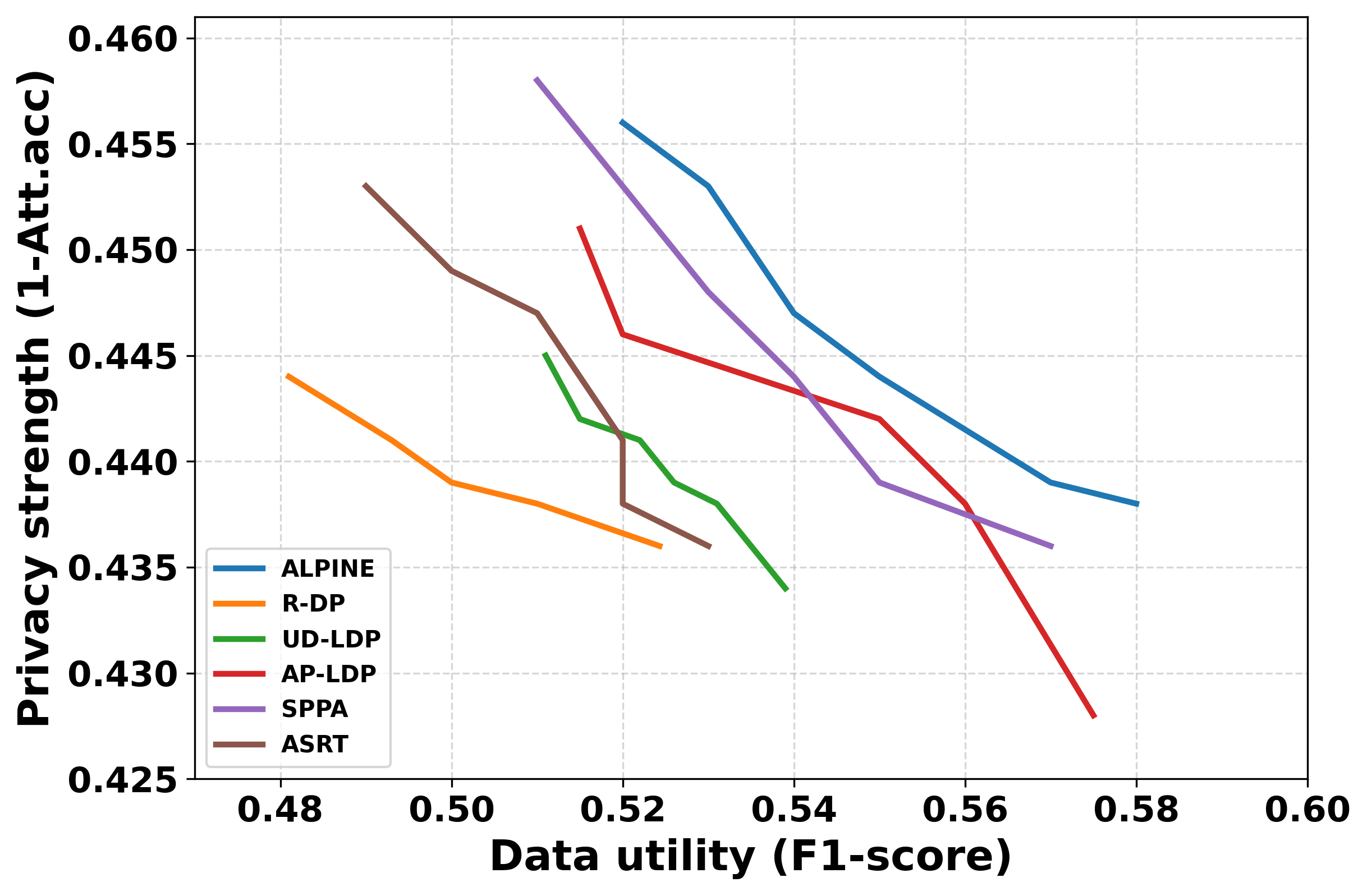} 
    \caption{Trade-off on Diabetes Dataset.}
    \label{fig:diabetes_privacy_utility}
  \end{minipage}
\end{figure*}

\subsubsection{Comparison with Dynamic and Adaptive Privacy-Preserving Methods}
To evaluate the effectiveness of ALPINE, we select several representative privacy-preserving methods in MECS as baselines. 
Specifically, \textbf{R-DP} \cite{shuai2022r} dynamically adjusts the protection strength through a closed-loop risk-awareness process together with a lightweight perturbation mechanism based on Bloom filters and data dissemination; 
\textbf{UD-LDP} \cite{10797681} achieves continuous data-stream privacy protection through w-adjacent-event privacy, entropy-driven privacy demand modeling, and adaptive window-based budget allocation; 
\textbf{AP-LDP} \cite{SONG2024103517} adaptively selects the perturbation mechanism between basic RAPPOR and k-RR based on the minimum mean squared error criterion; 
\textbf{ASRT} \cite{10678855} combines dynamic feature extraction, adaptive sampling, and a BLP mechanism to jointly preserve temporal patterns and privacy in finite-range time-series scenarios; 
\textbf{SPPA} \cite{10.1145/3696410.3714619} perturbs the sampling period and incorporates Fourier interpolation to preserve temporal relevance, thereby enabling locally differentially private release of infinite data streams.

To improve comparison fairness, we evaluate the privacy–utility trade-off of all methods under a unified protocol using the same dataset, the same downstream task, the same number of privacy-strength levels, and the same evaluation metrics. 
We compare the actual privacy protection effect and task utility achieved by each method at the corresponding operating points, rather than directly aligning their internal mechanisms. 
Utility is measured by F1-score on Intel and Diabetes, and by the classification performance of the freezer on/off prediction task on UK-DALE; 
privacy strength is measured by closeness to random guessing, attack-model reconstruction error, and reduction in attack success rate, respectively.

Figure 9–11 show that all methods exhibit privacy–utility conflicts to varying degrees. In contrast, ALPINE demonstrates a stronger overall trade-off on all three datasets, with its curves lying closer to the favorable region of the privacy–utility plane. This indicates that ALPINE can effectively suppress attack performance while maintaining strong task utility. The other baselines tend to be competitive only in limited operating regions; as utility improves, their privacy strength declines more rapidly, resulting in weaker overall robustness.

\begin{table}[h]
\centering
\caption{Communication overhead B/s on different devices.}
\label{tab:comm_overhead}
\begin{adjustbox}{max width=\linewidth}
\begin{tabular}{lcccccc}
\toprule
Device & ALPINE & R-DP & UD-LDP & AP-LDP & SPPA & ASRT \\
\midrule
Raspberry PI 5 & 64 & 167 & 300 & 100 & 70 & 72 \\
Portenta H7   & 92 & 200 & 360 & 132 & 102 & 104 \\
PC        & 60 & 165 & 300 & 96 & 68 & 71 \\
\bottomrule
\end{tabular}
\end{adjustbox}
\end{table}

\begin{table}[h]
\centering
\caption{Computational overhead ms on different devices.}
\label{tab:comp_overhead}
\begin{adjustbox}{max width=\linewidth}
\begin{tabular}{lcccccc}
\toprule
Device & ALPINE & R-DP & UD-LDP & AP-LDP & SPPA & ASRT \\
\midrule
Raspberry PI 5 & 42   & 55   & 32  & 8   & 7   & 3   \\
Portenta H7    & 52   & 60   & 40  & 10  & 10  & 5   \\
PC        & 12   & 14   & 7   & 0.5 & 0.4 & 0.5 \\
\bottomrule
\end{tabular}
\end{adjustbox}
\end{table}

To further evaluate deployment cost, we measure communication and computational overhead on Raspberry PI 5, Arduino Portenta H7, and PC. 
Communication overhead is measured in \textbf{B/s} as the average transmitted application-layer bytes per unit time, while computational overhead is measured in \textbf{ms} as the average latency of one round of device-side privacy processing. The results show that ALPINE achieves the lowest communication overhead on all three platforms, indicating that its lightweight closed-loop design effectively reduces redundant transmissions and privacy-related interaction costs. In terms of computational overhead, the average latency of ALPINE is marginally higher than that of AP-LDP, UD-LDP, SPPA, and ASRT, but lower than that of R-DP, suggesting that ALPINE does not simply pursue the minimum local computation delay; rather, it trades an acceptable computational cost for lower communication burden and stronger adaptive control capability. Overall, ALPINE achieves a more favorable trade-off at the system-cost level and is better suited for deployment in dynamic edge crowdsensing scenarios with limited bandwidth and constrained resources.

\begin{table}[h]
\centering
\small
\caption{Deployment of ALPINE on Edge Devices.}
\label{tab:edge-perf}
\begin{adjustbox}{max width=\linewidth}
\begin{tabular}{lcccc}
\toprule
Device &  Latency (s) & CPU (\%) & Memory (\%) & Power (W) \\
\midrule
Raspberry PI 5                 & 0.813 & 1.06 & 26.90 & 5.13 \\
Raspberry PI 5+ALPINE          & 0.934 & 2.40 & 29.30 & 5.45 \\
Portenta H7      & 0.777   &18.80  &  5.01  & 0.64 \\
Portenta H7+ALPINE & 0.857 & 20.90  & 7.50   & 0.82 \\
\bottomrule
\end{tabular}
\end{adjustbox}
\end{table}

\subsubsection{Large-Scale Experimental Deployment}
We deploy ALPINE on two representative terminal devices, Raspberry PI 5 and Arduino Portenta H7, to collect temperature readings every 2 seconds and stream them to an edge server in real time. We define system latency as the wall-clock time from the onset of sensor-data acquisition to the completion of on-device privacy processing and subsequent transmission to the server via MQTT. 
As Table \ref{tab:edge-perf} shows, despite online LightAE selection and dynamic noise injection, the additional processing delay introduced by ALPINE remains modest. Meanwhile, CPU utilization remains low overall, and the increases in memory footprint and energy consumption are moderate.
\begin{figure}[h]
  \centering
  \includegraphics[width=0.7\linewidth]{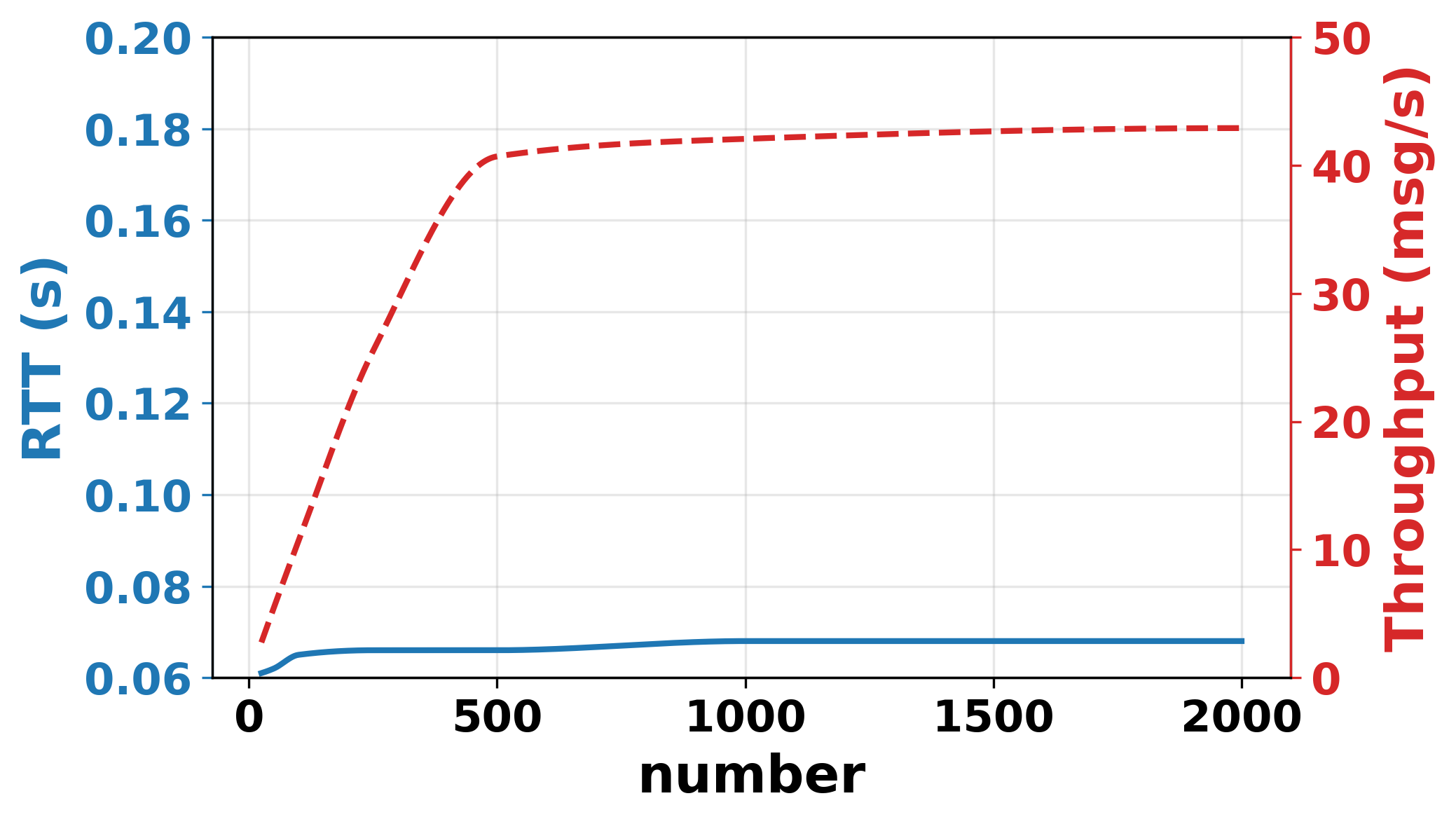}
  \caption{Large-Scale Deployment of ALPINE.}
  \label{fig:lss}
\end{figure}

We also use five Raspberry PI devices to run varying numbers of concurrent processes, emulating the ingress of a large population of terminal devices. The server performs a lightweight classification and returns an acknowledgment, while terminals log the round-trip time (RTT) for each message and the server measures throughput. 
In Figure 12, latency remains stable even under high concurrency, indicating that the computational overhead of ALPINE is well controlled. With thousands of terminals, throughput saturates, suggesting the need for further optimization under extreme concurrency. Overall, the ALPINE system demonstrates the feasibility of large-scale edge deployment and promising scalability.

\section{Conclusion}

We propose \textbf{ALPINE}, a lightweight closed-loop adaptive privacy budget allocation framework for MECS. 
It performs on-device multi-dimensional risk perception and uses an offline-trained TD3 policy to allocate privacy budgets under dynamic constraints. 
The edge server evaluates privacy strength and downstream utility, and feeds back signals for online policy switching and periodic offline refinement. 
Experiments on real-world and prototype deployment datasets show that ALPINE achieves a favorable balance among anomaly detection, privacy–utility trade-off, efficiency, and robustness. 
In future work, we will explore lighter anomaly detectors and CMDP-based control with primal–dual optimization to further improve principled online adaptation.

\bibliographystyle{IEEEtran}
\bibliography{refs}

\begin{table*}[t]
\centering
\footnotesize
\setlength{\tabcolsep}{3pt}
\renewcommand{\arraystretch}{1.0}
\caption{Comparison of representative privacy-preserving frameworks for MECS.}
\label{tab:compare_mecs_privacy}
\begin{tabular}{C{1.2cm} C{1.9cm} C{2.0cm} C{1.8cm} C{2.2cm} C{2.0cm} C{1.5cm} C{1.9cm}}
\toprule
\textbf{Category} &
\textbf{Method} &
\textbf{Objective} &
\textbf{Adaptation} &
\textbf{Decision} &
\textbf{Mechanism} &
\textbf{Feedback} &
\textbf{Deployability} \\
\midrule
 Static&  \cite{ZHANG2024103464} & Location privacy & Fixed & Fixed-budget DP perturbation & Laplace DP & None & On-device \\
 \midrule
 Static & \cite{10945149}       &  Efficient selection under location privacy & Fixed & Matrix-based matching & $k$-anonymity & None & Cloud-assisted \\
\midrule
Collaborative & \cite{11045510} & Utility-privacy trade-off & Task-driven & Client-server protocol  & Secure computation / private inference & None & Vision + secure comp \\
\midrule
Collaborative & \cite{11145334} & Trust and conditional anonymity & Fog–edge cooperative & Cryptographic protocol & Conditional privacy + invalid-data filtering & Cloud-side monitoring &  Fog-assisted; crypto overhead \\
\midrule

Adaptive & \cite{shuai2022r} & Utility-privacy trade-off & External risk + adversary model & Rule-based mapping & Bloom filter +  perturbation &  Malicious-user detection & Low communication cost \\
\midrule
Adaptive & \cite{CHEN2024110613} & Task quality maximization & Per-task allocation &  Quality-aware online optimization & Homomorphic encryption & None & Crypto overhead\\
\midrule
Adaptive & \cite{10963891} & Utility, privacy, and credibility & Per-round trust update & Combinatorial MAB & Truth discovery + verification & Short-/long-term trust validation & UAV-assisted; multi-module overhead \\
\midrule
\textbf{Adaptive} & \textbf{ALPINE} & \textbf{ Privacy, utility, and energy} & \textbf{Channel/data quality/device state} & \textbf{Offline TD3 policy set + online switching} & \textbf{ BLP noise injection} & \textbf{Edge-side utility and leakage assessment} & \textbf{On-device inference} \\
\bottomrule
\end{tabular}
\end{table*}

\section{Appendix}

\subsection{Detailed Comparison with Prior Privacy Frameworks}

Table~\ref{tab:compare_mecs_privacy} provides a structured comparison of representative MECS privacy-protection frameworks across six dimensions. 
The comparison underscores that existing static approaches typically lack runtime adaptation, while collaborative designs often incur non-trivial communication/cryptographic overhead and still provide limited fine-grained control. 

\subsection{Proof of Lemma 1}

Our reward function in TD3 is in (5) and (6). 
We assume the $\text {EnergyCost}$ is independent of variations in the privacy budget.
Thus, to formalize the relationship between the $\epsilon$ and the reward function $W$, we express it as follows:
$
\mathrm{R}(\varepsilon)=\alpha S(s) U(\varepsilon)-\beta P(s) V(\varepsilon)-E
$,$E$ is a constant, where:
\begin{equation}
\begin{aligned}
S(s)&=\frac{1}{1+\exp \left(-k\left(s-s_{0}\right)\right)}, \quad U(\varepsilon)=\left(\frac{\varepsilon_{\max }-\varepsilon}{\varepsilon_{\max }-\varepsilon_{\min }}\right)^{\delta},\\
P(s)&=1-\rho s>0, \quad V(\varepsilon)=\left(\frac{\sigma_{0}}{\varepsilon}\right)^{2}, 0<\delta<1,\sigma_{0}>0.
\label{eq:isolation-qweqe
}
\end{aligned}
\end{equation}

Taking the derivative of $R(\varepsilon)$  with respect to $\epsilon$, we obtain:
\begin{equation}
\frac{d \mathrm{R}}{d \varepsilon}=-\alpha S(s) \cdot \frac{\delta}{\varepsilon_{\max }-\varepsilon_{\min }}\left(\frac{\varepsilon_{\max }-\varepsilon}{\varepsilon_{\max }-\varepsilon_{\min }}\right)^{\delta-1}+\beta P(s) \cdot \frac{2 \sigma_{0}^{2}}{\varepsilon^{3}} .
\label{eq:isolation-qweqe
}
\end{equation}

$0<\delta<1$. As $\varepsilon \rightarrow \varepsilon_{\min }^{+}$ in (11), both the privacy derivative and the utility derivative are finite.

\begin{equation}
\frac{d \mathrm{R}}{d \varepsilon}=-\frac{\alpha S(s) \delta}{\left(\varepsilon_{\max }-\varepsilon_{\min }\right)}+\beta P(s) \cdot \frac{2 \sigma_{0}^{2}}{\varepsilon_{\min }^{3}} .
\label{eq:isolation-qqq}
\end{equation}
$\varepsilon \rightarrow \varepsilon_{\max }^{-}$: $\left(\frac{\varepsilon_{\max }-\varepsilon}{\varepsilon_{\max }-\varepsilon_{\min }}\right)^{\delta-1} \rightarrow+\infty$, therefore, $\frac{d R}{d \varepsilon} \rightarrow-\infty$. 

The second derivative is given by:
\begin{equation}
\frac{d^{2} \mathrm{R}}{d \varepsilon^{2}}=\frac{\alpha S(s) \delta(\delta-1)}{\left(\varepsilon_{\max }-\varepsilon_{\min }\right)^{\delta}}\left(\varepsilon_{\max }-\varepsilon\right)^{\delta-2}-\beta P(s) \cdot \frac{6 \sigma_{0}^{2}}{\varepsilon^{4}}
\label{eq:isolation-rrr}.
\end{equation}

When $0<\delta<1$, it is evident that $\frac{d^{2} R}{d \varepsilon^{2}}<0$, indicating the reward function is strictly concave over the interval.

Moreover, $\frac{d R}{d \varepsilon}$ is continuous and decreasing. 
If $\left.\frac{d R}{d \varepsilon}\right|_{\varepsilon \rightarrow \varepsilon_{\min }^{+}}>0$, namely $\beta P(s) \cdot 2 \sigma_{0}^{2} / \varepsilon_{\min }^{3}>\alpha S(s) \delta /\left(\varepsilon_{\max }-\varepsilon_{\min }\right)$, by the intermediate value theorem and the strict concavity, there exists a unique $\varepsilon^{*} \in\left(\varepsilon_{\min }, \varepsilon_{\max }\right)$ such that $\frac{d R}{d \varepsilon}=0,$ corresponding to the unique global maximum. 
If $\left.\frac{d R}{d \varepsilon}\right|_{\varepsilon \rightarrow \varepsilon_{\min }^{+}}<0$, since the derivative is strictly decreasing and approaches negative infinity as $\epsilon\rightarrow\varepsilon_{\max }$, the function strictly decreases over the entire interval. Therefore, the global maximum occurs at the boundary $\varepsilon_{\min }$.

Therefore, it is necessary to adjust the parameters to satisfy $\left.\frac{d R}{d \varepsilon}\right|_{\varepsilon \rightarrow \varepsilon_{\min }^{+}}>0$. 
Under the condition $0<\delta<1$ and a reasonably bounded setting, the function $R(\varepsilon)$ is concave within the interval $\left(\varepsilon_{\min }, \varepsilon_{\max }\right)$, and its first derivative is strictly decreasing with opposite signs at the interval boundaries. 

For $\varepsilon^{*}$, it satisfies $\frac{d R}{d \varepsilon}=0$. This condition corresponds to the first-order optimality condition in multi-objective optimization under the Karush-Kuhn-Tucker (KKT) framework \cite{ghojogh2021kkt}, indicating that the optimal budget point $\varepsilon^{*}(s)$ is achieved when the marginal privacy gain equals the marginal utility loss, weighted by their respective trade-off coefficients:
\begin{equation}
\alpha \cdot \frac{\partial \text { PrivacyGain }}{\partial \varepsilon}=\beta \cdot \frac{\partial \text { UtilityLoss }}{\partial \varepsilon}
\label{eq:isolation-ttt}.
\end{equation}

\end{document}